\documentclass[final]{iccv}

\usepackage{times}
\usepackage{epsfig}
\usepackage{graphicx}
\usepackage{amsmath}
\usepackage{amssymb}
\usepackage{multirow}
\usepackage{pifont}
\usepackage{booktabs}
\usepackage{epsfig}
\usepackage{booktabs,multirow}
\usepackage{graphics}
\usepackage{threeparttable}
\usepackage{color}
\usepackage[normalem]{ulem}
\usepackage{multirow}
\usepackage{float}
\usepackage{amsfonts}
\usepackage{bm}
\usepackage{enumitem}
\usepackage[numbers]{natbib}
\usepackage{array}
\usepackage[table]{xcolor}
\usepackage{colortbl}
\definecolor{myy}{RGB}{126,95,0}
\definecolor{mygray}{gray}{.9}
\definecolor{bblue}{RGB}{30,80,120}
\definecolor{mygray1}{gray}{.7}
\usepackage{bm}
\usepackage{mathtools}
\usepackage{subcaption}
\usepackage{siunitx}
\usepackage[nopar]{lipsum}
\usepackage[export]{adjustbox}

\newcolumntype{I}{!{\vrule width 1pt}}

\definecolor{ggray}{RGB}{127,127,127}
\definecolor{mygreen}{RGB}{93,174,86}

\makeatletter
\newcommand{\thickhline}{%
	\noalign {\ifnum 0=`}\fi \hrule height 1pt
	\futurelet \reserved@a \@xhline
}
\makeatother
\newcommand{\tabincell}[2]{\begin{tabular}{@{}#1@{}}#2\end{tabular}}

\usepackage{caption}
\usepackage[pagebackref=true,breaklinks=true,colorlinks,bookmarks=false]{hyperref}
\usepackage[utf8]{inputenc}

\usepackage{cleveref}
\crefname{section}{§}{§§}
\Crefname{section}{§}{§§}

\def\iccvPaperID{6230} 
\def\confYear{ICCV 2021}
\begin{document}



\title{Exploring Cross-Image Pixel Contrast for Semantic Segmentation}

\author{Wenguan Wang$^{1}\thanks{The first two authors contribute equally to this work.}$~,~~\hspace{1pt}Tianfei Zhou$^{1*}$,~~Fisher Yu$^{1}$~,~~Jifeng Dai$^{2}$,~~Ender Konukoglu$^{1}$,~~Luc Van Gool$^{1}$   \\
	{$^1$}  Computer Vision Lab, ETH Zurich \hspace{0pt}
	{$^2$}  SenseTime Research \hspace{0pt}
	%
}

\maketitle

\begin{abstract}
Current semantic segmentation methods focus only on mining ``local'' context, \ie, dependencies between pixels within individual images, by context-aggregation modules (\eg, dilated convolution, neural attention) or structure-aware$_{\!}$ optimization$_{\!}$ criteria$_{\!}$ (\eg, IoU-like$_{\!}$ loss).$_{\!}$ However, they ignore ``global''~context of the training data, \ie, rich semantic relations between pixels across different images. Inspired by recent advance in unsupervised contrastive representation learning, we propose a pixel-wise contrastive algorithm for semantic segmentation in the fully supervised setting. The core~idea~is~to~enforce pixel embeddings belonging to a same semantic class~to~be~more similar than embeddings from different classes.~It raises a {pixel-wise metric learning paradigm for semantic segmentation, by explicitly exploring the structures of labeled pixels}, which were rarely explored before. Our method~can~be effortlessly incorporated into existing segmentation frameworks without extra overhead during testing. We  experimentally show that, with famous segmentation models$_{\!}$ (\ie,$_{\!}$ DeepLabV3,$_{\!}$ HRNet,$_{\!}$ OCR)  and  backbones$_{\!}$  (\ie, ResNet, HRNet), our method brings performance improvements across diverse datasets (\ie, Cityscapes, PASCAL-Context, COCO-Stuff, CamVid). We expect this work will encourage our community to rethink the current de facto training paradigm in semantic segmentation.\footnote{Our code will be available at \url{https://github.com/tfzhou/ContrastiveSeg}.}


\end{abstract}

\vspace{-6pt}
\section{Introduction}
Semantic segmentation, which aims to infer semantic labels for all pixels in an image, is a fundamental problem in computer vision. In the last decade, semantic segmentation has achieved remarkable progress,  driven by the availability of large-scale datasets (\eg, Cityscapes~\cite{cordts2016cityscapes}) and rapid evolution of convolutional networks (\eg, VGG~\cite{Simonyan15}, ResNet~\cite{he2016deep}) as well as segmentation models (\eg, fully convolutional network (FCN)~\cite{long2015fully}). In particular, FCN~\cite{long2015fully} is the cornerstone of modern deep learning techniques for segmentation, due to its unique advantage in end-to-end pixel-wise representation learning. However, its spatial invariance nature hinders the ability of modeling useful context among pixels (within images). 
Thus a main stream of subsequent effort delves into network designs for effective context
aggregation, \eg, dilated convolution\!~\cite{yu2015multi,chen2017deeplab,chen2017rethinking}, spatial pyramid pooling\!~\cite{zhao2017pyramid}, multi-layer feature fusion\!~\cite{ronneberger2015u,lin2017refinenet} and neural attention\!~\cite{hu2018squeeze,fu2019dual}. In addition, as the widely adopted pixel-wise cross entropy loss fundamentally lacks the spatial discrimination power, some alternative optimization criteria are proposed to explicitly address object structures during segmentation network training~\cite{ke2018adaptive,berman2018lovasz,2019_zhao_rmi}.

\begin{figure}[t]
	\begin{center}
		\includegraphics[width=\linewidth]{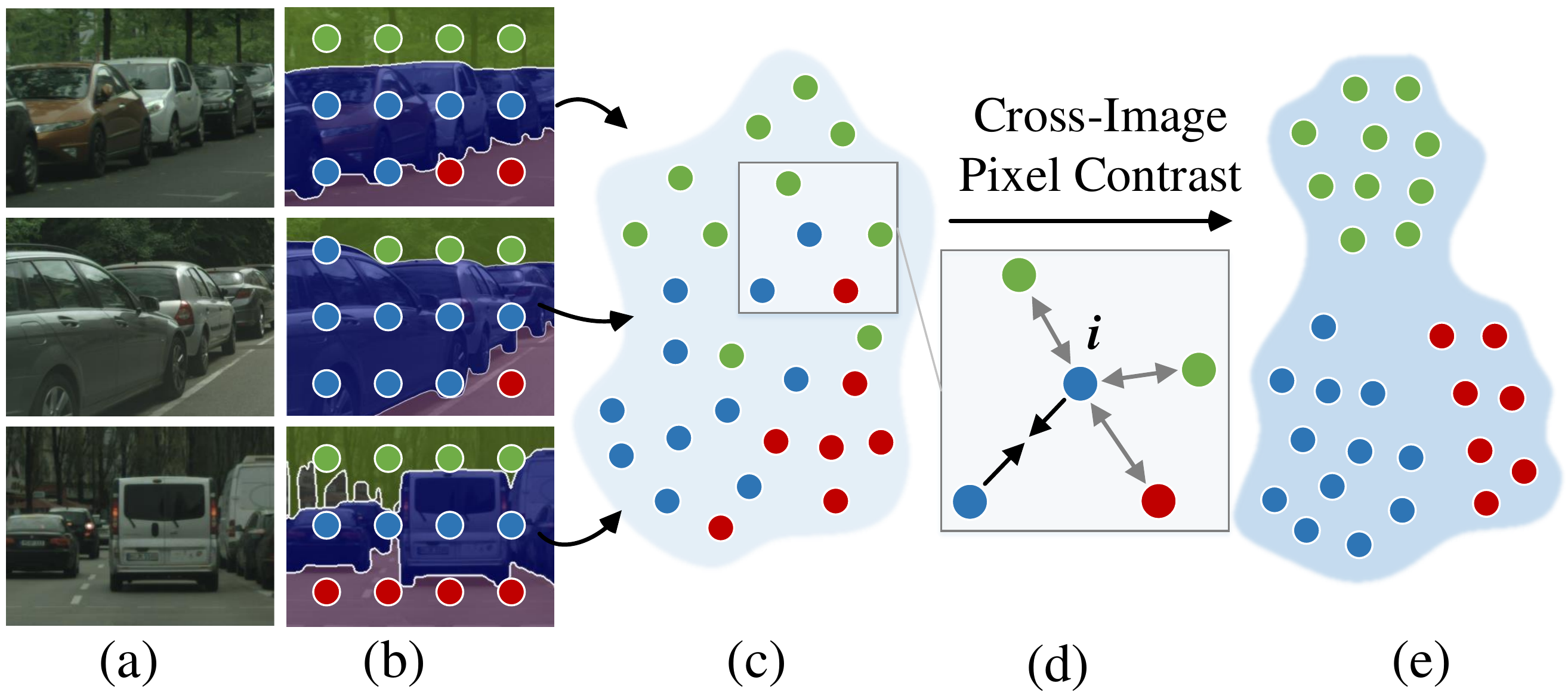}
	\end{center}
	\vspace{-18pt}
	\captionsetup{font=small}
	\caption{\small\!\textbf{Main$_{\!}$ idea.}$_{\!}$ Current$_{\!}$ segmentation$_{\!}$ models$_{\!}$ learn to map pixels (b) to an embedding space (c), yet ignoring intrinsic structures of labeled data (\ie, inter-image relations among pixels from a same class, noted with same color in~\!(b)). Pixel-wise contrastive learning is introduced to foster a new training paradigm (d), by \textit{explicitly} addressing intra-class compactness and inter-class dispersion. Each pixel (embedding) $i$ is pulled closer (\protect\includegraphics[scale=0.01,valign=c]{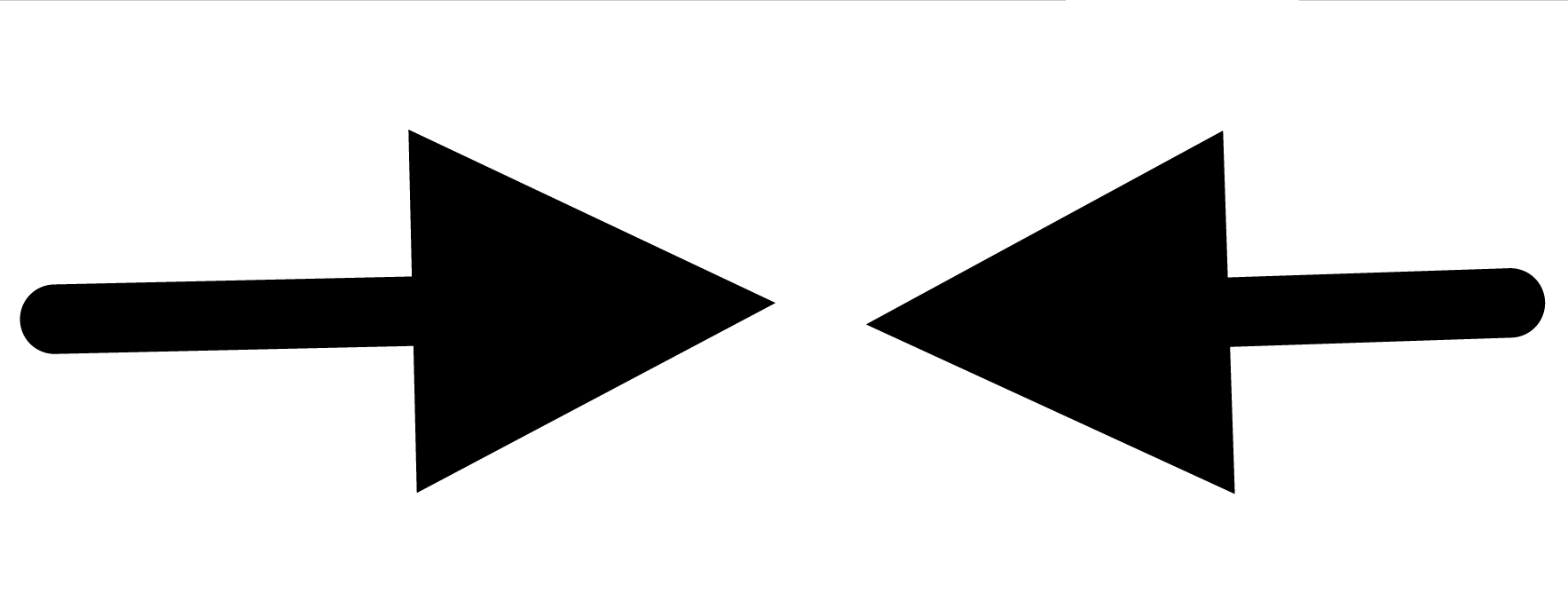}) to~pixels~(\protect\includegraphics[scale=0.008,valign=c]{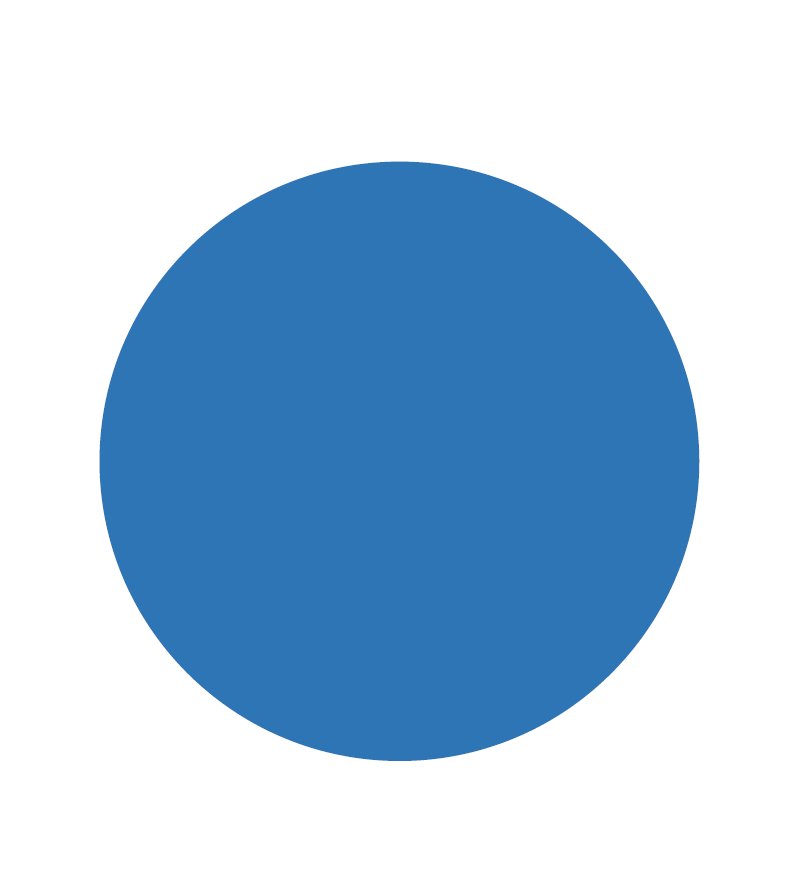}) of the same class, but pushed far (\protect\includegraphics[scale=0.01,valign=c]{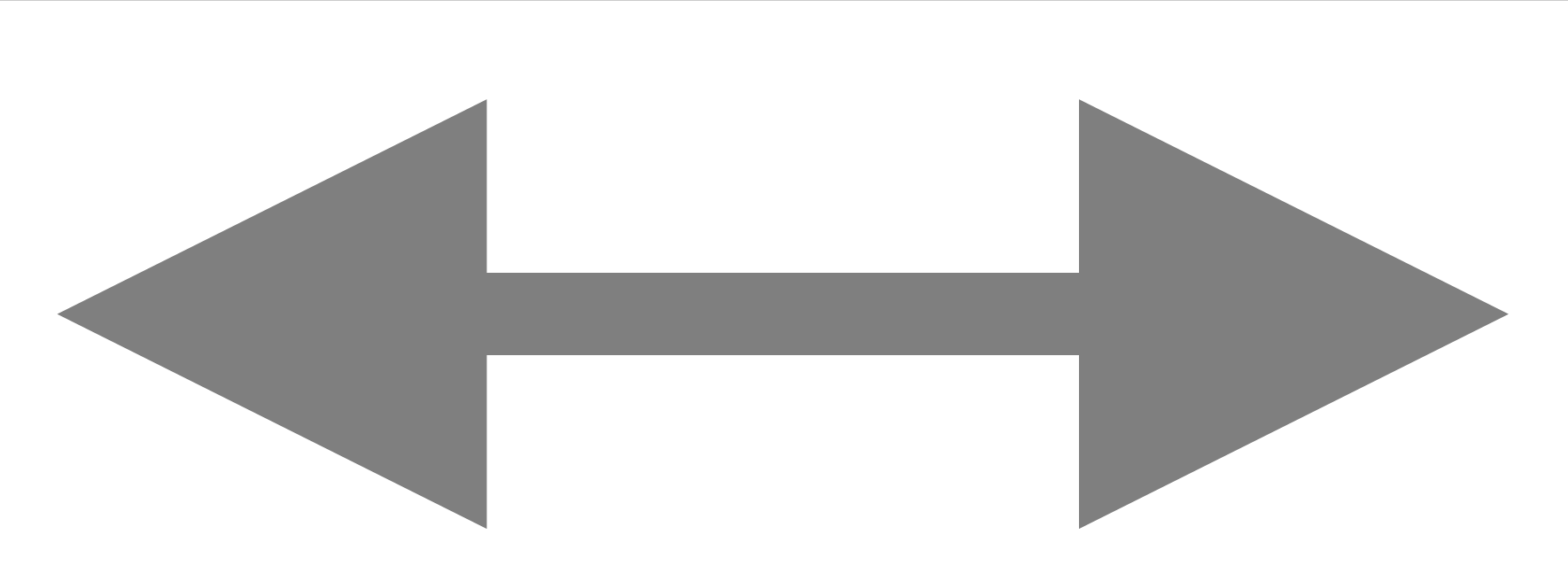}) from pixels (\protect\includegraphics[scale=0.008,valign=c]{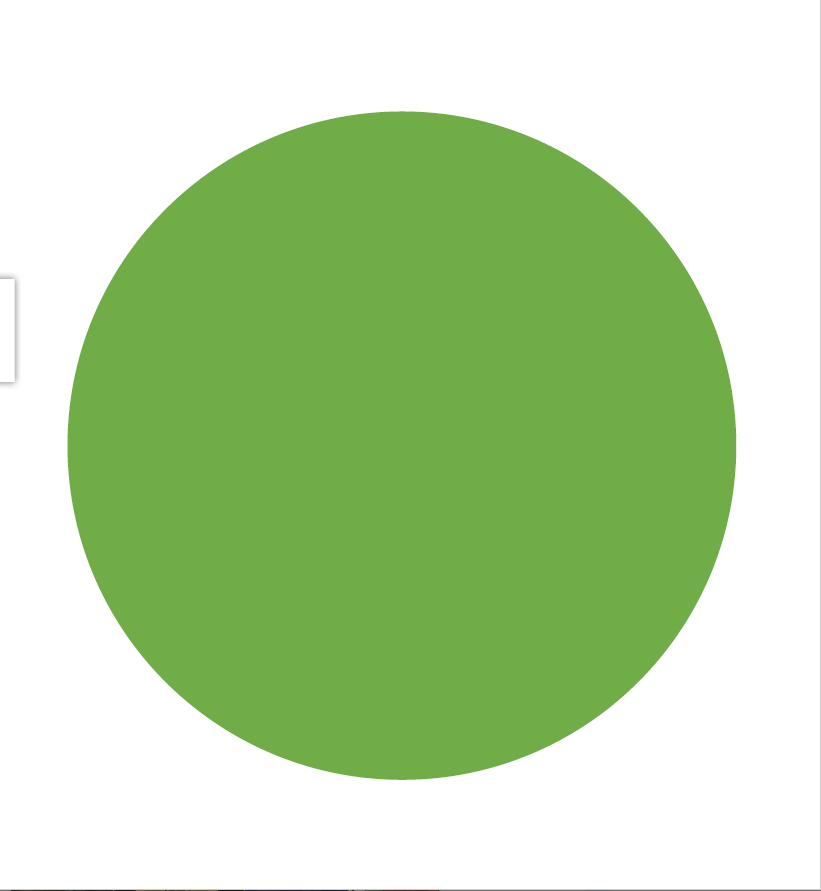} \protect\includegraphics[scale=0.008,valign=c]{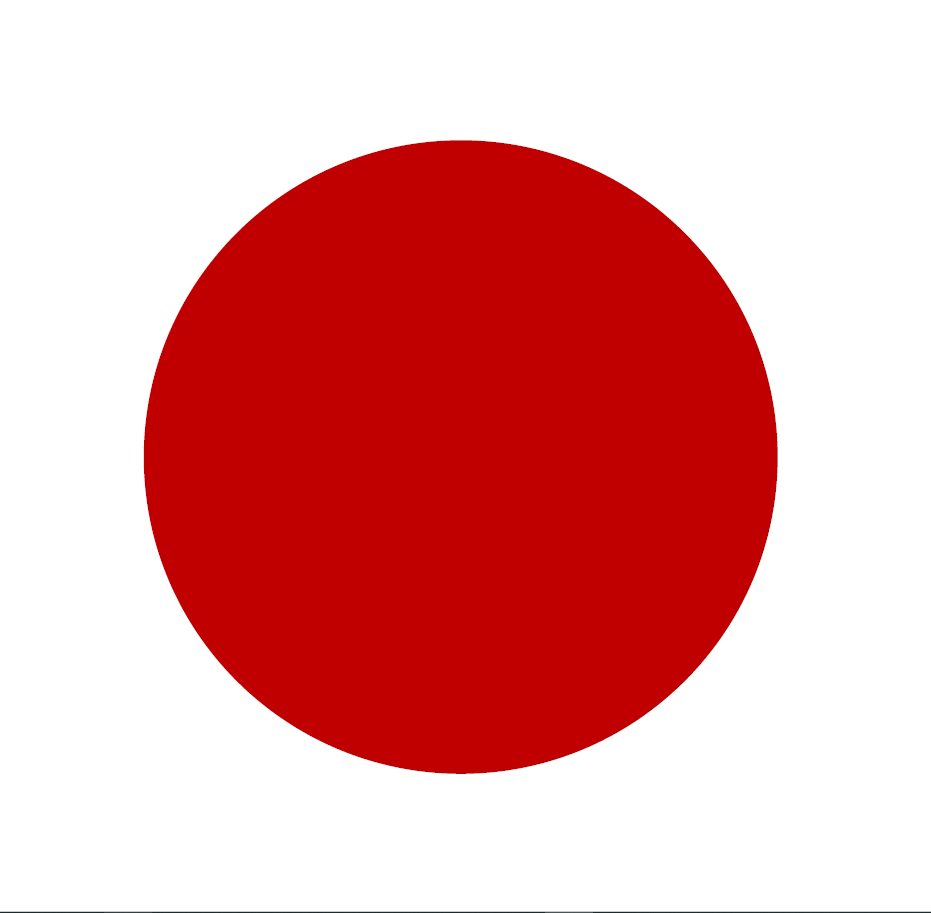}) from other$_{\!}$ classes.$_{\!}$ Thus$_{\!}$ a$_{\!}$ better-structured embedding$_{\!}$ space~(e)~is~derived, eventually$_{\!}$ boosting the$_{\!}$ performance$_{\!}$ of$_{\!}$ segmentation$_{\!}$ models.$_{\!\!}$}
	\vspace{-13pt}
	\label{fig:motivation}
\end{figure}


Basically, these segmentation models (excepting \cite{hwang2019segsort}) utilize deep architectures to project image pixels into a highly non-linear embedding space (Fig.\!~\ref{fig:motivation}(c)). However, they typically learn the embedding space that only makes use of ``local'' context around pixel samples (\ie, pixel dependencies \textit{within} individual images), but ignores ``global'' context of the whole dataset (\ie, pixel semantic relations \textit{across} images). Hence, an essential issue has been long ignored in the field:  \textit{what should a good segmentation embedding space look like}? Ideally, it should not only \textbf{1)} address the categorization ability of individual pixel embeddings, but~also$_{\!}$ \textbf{2)}$_{\!}$ be$_{\!}$ well$_{\!}$ structured$_{\!}$ to$_{\!}$ address intra-class compactness and inter-class dispersion. With regard to \textbf{2)}, pixels from a same class should be closer than those from different classes, in the embedding space.  Prior studies\!~\cite{liu2016large,schroff2015facenet} in representation learning also suggested that encoding intrinsic structures of training data (\ie, \textbf{2)}) would  facilitate feature discriminativeness (\ie, \textbf{1)}).  So we speculate that, although existing algorithms have achieved impressive performance, it is possible to learn a better structured pixel embedding space by considering both \textbf{1)} and \textbf{2)}.


Recent advance in unsupervised representation learning~\cite{chen2020simple,he2020momentum} can be ascribed to the resurgence of contrastive learning -- an essential branch of deep metric learning~\cite{kaya2019deep}. The core idea is ``learn to compare'':  given an \textit{anchor} point, distinguish a similar (or \textit{positive}) sample from a set of dissimilar (or \textit{negative}) samples,  in a projected embedding space. Especially, in the field of computer vision, the contrast is evaluated based on image feature vectors; the augmented version of an anchor image is viewed as a positive, while all the other images in the dataset act as negatives.

The great success of unsupervised contrastive learning and our aforementioned speculation together motivate us to rethink the current de facto training paradigm in semantic segmentation. Basically, the power of unsupervised contrastive learning roots from the structured comparison loss, which takes the advantage of the context within the training data. With this insight, we propose a pixel-wise contrastive algorithm for more effective dense representation learning in the \textbf{fully supervised setting}. Specifically, in addition to adopting the pixel-wise cross entropy loss to address class discrimination (\ie, property \textbf{1)}), we utilize a pixel-wise contrastive loss to further shape the pixel embedding space, through exploring the structural information of labeled pixel samples (\ie, property \textbf{2)}). The idea of the pixel-wise contrastive loss is to compute \textit{pixel-to-pixel contrast}: enforce embeddings to be similar for positive pixels, and dissimilar for negative ones. As the pixel-level categorical information is given during training, the positive samples are the pixels belonging to a same class, and the negatives are the pixels from different classes (Fig.\!~\ref{fig:motivation}(d)). In this way, the global property of the embedding space can be captured (Fig.\!~\ref{fig:motivation}(e)) for better reflecting intrinsic structures of training data and enabling more accurate segmentation predictions.

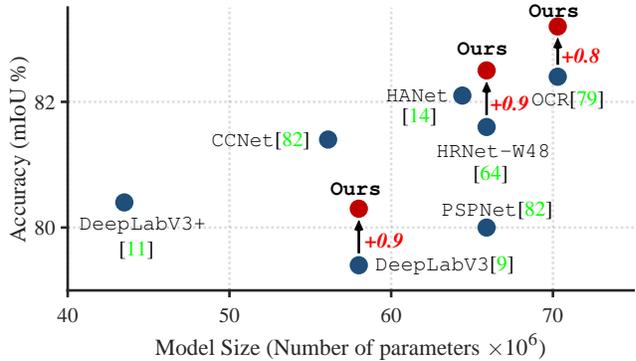
\begin{figure}[t]
	\vspace{-2pt}
	\begin{center}
		\includegraphics[width=\linewidth]{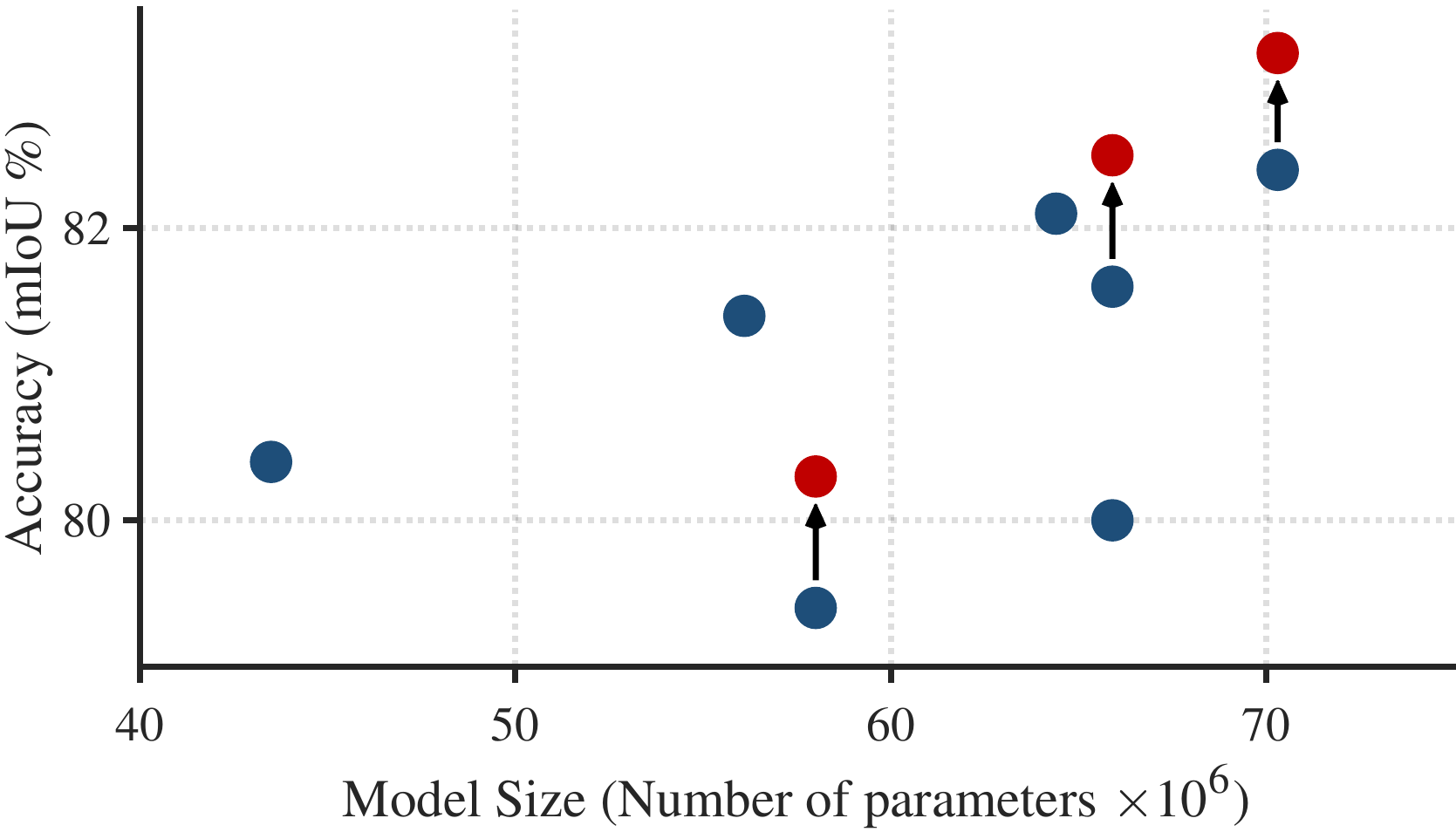}
		\put(-75,77.5){\footnotesize $\texttt{HRNet-W48}$}
		\put(-61,68.5){\footnotesize \cite{wang2020deep}}
		\put(-210,50){\footnotesize $\texttt{DeepLabV3+}$}
		\put(-195,40){\footnotesize \cite{chen2018encoder}}
		\put(-98,34.5){\footnotesize $\texttt{DeepLabV3}$\cite{chen2017rethinking}}
		\put(-39,97){\footnotesize $\texttt{OCR}$\cite{yuan2020object}}
		\put(-73,55){\footnotesize $\texttt{PSPNet}$\cite{zhao2017pyramid} }
		\put(-160,81.5){\footnotesize $\texttt{CCNet}$\cite{zhao2017pyramid} }
		\put(-94,98.5){\footnotesize $\texttt{HANet}$}
		\put(-88,90.5){\footnotesize \cite{choi2020cars}}
		\put(-115,63){\footnotesize $\textbf{\texttt{Ours}}$}
		\put(-67,116){\footnotesize $\textbf{\texttt{Ours}}$}
		\put(-40,130.5){\footnotesize $\textbf{\texttt{Ours}}$}
		\put(-27.5,113){\footnotesize \color{red}{$\textbf{\textit{+0.8}}$}}
		\put(-54.5,95){\footnotesize \color{red}{$\textbf{\textit{+0.9}}$}}
		\put(-102,43){\footnotesize \color{red}{$\textbf{\textit{+0.9}}$}}
	\end{center}
	\vspace{-18pt}
	\captionsetup{font=small}
	\caption{\small\!\!\!\textbf{Accuracy~\!\textit{vs}.~\!model~\!size} on Cityscapes~\!\texttt{test}\!~\cite{cordts2016cityscapes}.~\!Our contrastive enables consistent performance improvements over state-of-the-arts, \ie, DeepLabV3~\!\cite{chen2017rethinking}, HRNet~\!\cite{wang2020deep}, OCR~\!\cite{yuan2020object}, without bringing any change to base networks during inference.}
	\vspace{-13pt}
	\label{fig:improve}
\end{figure}

With our supervised pixel-wise contrastive algorithm, two novel techniques are developed. \textbf{First}, we propose a \textit{region} memory bank to better address the nature of semantic segmentation. Faced with huge amounts of highly-structured pixel training samples, we let the memory store pooled features of semantic regions (\ie, pixels with a same semantic label from a same image), instead of pixel-wise embeddings only. This leads to \textit{pixel-to-region contrast}, as a complementary for the pixel-to-pixel contrast strategy. Such memory design allows us to access more representative data samples during each training step and fully explore structural relations between pixels and semantic-level segments, \ie, pixels and segments belonging to a same class should be close in the embedding space. \textbf{Second}, we propose different sampling strategies to make better use of informative samples and let the segmentation model pay more attention to those segmentation-hard pixels. Previous works have confirmed that hard negatives are crucial for metric learning~\cite{kaya2019deep,schroff2015facenet,simo2015discriminative}, and our study further reveals the importance of mining both informative negatives/positives and anchors in this supervised, dense image prediction task.

In a nutshell, our \textbf{contributions} are three-fold:
\begin{itemize}[leftmargin=*]
	\setlength{\itemsep}{0pt}
	\setlength{\parsep}{-2pt}
	\setlength{\parskip}{-0pt}
	\setlength{\leftmargin}{-15pt}
	\vspace{-7pt}
\item We propose a supervised, pixel-wise contrastive learning method for semantic segmentation. It lifts current image-wise training strategy to an inter-image, pixel-to-pixel paradigm. It  essentially learns a \textit{well structured} pixel semantic embedding space, by making full use of the global semantic similarities among labeled pixels.
\item We develop a region memory to better explore the large visual data space and support to further calculate pixel-to-region contrast. Integrated with pixel-to-pixel contrast computation, our method exploits semantic correlations among pixels, and between pixels and semantic regions.
\item We demonstrate that more powerful segmentation models with better example and anchor sampling strategies could be delivered instead of selecting random pixel samples.
\vspace*{-6pt}
\end{itemize}



Our method can be seamlessly incorporated into existing segmentation networks without any changes to the base model and without extra inference burden during testing (Fig.~\ref{fig:improve}). Hence, our method shows consistently improved intersection-over-union segmentation scores over challenging datasets (\ie, Cityscapes\!~\cite{cordts2016cityscapes}, PASCAL-Context\!~\cite{mottaghi2014role}, COCO-Stuff\!~\cite{caesar2018coco} and CamVid\!~\cite{brostow2009semantic}), using state-of-the-art segmentation architectures (\ie, DeepLabV3\!~\cite{chen2017rethinking}, HRNet\!~\cite{wang2020deep} and OCR\!~\cite{yuan2020object}) and famous backbones (\ie, ResNet\!~\cite{he2016deep}, HRNet\!~\cite{wang2020deep}). The impressive results shed light on the promises of metric learning in dense image prediction tasks. We expect this work to provide insights into the critical role of global pixel relationships in segmentation network training, and foster research on the open issues raised.





\vspace{-4pt}
\section{Related Work}%
Our work draws on existing literature in semantic segmentation, contrastive learning and deep metric learning. For brevity, only the most relevant works are discussed.

\noindent\textbf{Semantic Segmentation}. FCN~\cite{long2015fully} greatly promotes the advance of semantic segmentation. It is good at end-to-end dense feature learning, however, only perceiving limited visual context with local receptive fields. As strong dependencies exist among pixels in an image and these dependencies are informative about the  structures of the objects~\cite{wang2004image}, how to capture such dependencies becomes a vital issue for further improving FCN. A main group of follow-up effort attempts to aggregate multiple pixels to explicitly model context, for example, utilizing different sizes of convolutional/pooling kernels or dilation rates to gather multi-scale visual cues~\cite{yu2015multi,zhao2017pyramid,chen2017deeplab,chen2017rethinking}, building image pyramids to extract context from multi-resolution inputs, adopting the Encoder-Decoder
architecture to merge features from different network layers~\cite{ronneberger2015u,lin2017refinenet}, applying CRF to recover detailed
structures~\cite{liu2017deep,zheng2015conditional}, and employing neural attention~\cite{wang2018non} to directly exchange context between paired pixels~\cite{chen2016attention,hu2018squeeze,huang2019ccnet,fu2019dual}. Apart from investigating context-aggregation network modules, another line of work turns to designing context-aware optimization objectives~\cite{ke2018adaptive,berman2018lovasz,2019_zhao_rmi}, \ie, directly verify segmentation structures during training, to replace the pixel-wise cross entropy loss.

Though impressive, these methods only address pixel dependencies within individual images, neglecting the global context of the labeled data, 
\ie, pixel semantic correlations across different training images. Through a pixel-wise contrastive learning formulation, we map pixels in different categories to more distinctive features. The learned pixel features are not only discriminative for semantic classification within images, but also, more crucially, across images.

\noindent\textbf{Contrastive Learning.}
Recently, the most compelling methods for learning representations without labels have been unsupervised contrastive learning~\cite{oord2018representation,hjelm2019learning,wu2018unsupervised,chen2020improved,chen2020simple}, which significantly outperformed other pretext task-based alternatives~\cite{larsson2016learning,gidaris2018unsupervised,doersch2015unsupervised,noroozi2016unsupervised}. With a similar idea to exemplar learning~\cite{dosovitskiy2014discriminative}, contrastive methods learn representations in a discriminative manner by contrasting similar (positive) data pairs against dissimilar (negative) pairs. A major branch of subsequent studies focuses on how to select the positive and negative pairs. For image data, the standard positive pair sampling strategy is to apply strong perturbations to create multiple views of each image data~\cite{wu2018unsupervised,chen2020simple,he2020momentum,hjelm2019learning,caron2020unsupervised}. Negative pairs are usually randomly sampled, but some hard negative example mining strategies~\cite{khosla2020supervised,robinson2020contrastive,kalantidis2020hard} were recently proposed. In addition, to store more negative samples during contrast computation, fixed~\cite{wu2018unsupervised} or momentum updated~\cite{misra2020self,he2020momentum} memories are adopted. Some latest studies \cite{khosla2020supervised,henaff2020data,wei2020can} also confirm label information can assist contrastive learning based image-level pattern pre-training.

We raise a \textit{pixel-to-pixel} contrastive learning method for semantic segmentation in the fully supervised setting. It yields a new training protocol that explores global pixel$_{\!}$ relations in labeled data for regularizing segmentation embedding space. Though a few concurrent~works~also address contrastive learning in dense image prediction\!~\cite{xie2020propagate,chaitanya2020contrastive,wang2020dense}, the ideas are significantly different. First, they typically consider contrastive learning as a \textit{pre-training} step for dense image embedding. Second, they simply use the local context within individual images, \ie, only compute the contrast among pixels from augmented versions of a \textit{same} image.  Third, they do not notice the critical role of metric learning in complementing current well-established pixel-wise cross-entropy loss based training regime (\textit{cf.}$_{\!}$ \S\ref{sec:pc}).




\noindent\textbf{Deep Metric Learning.} The goal of metric learning is to quantify the similarity  among samples using an optimal distance metric. Contrastive loss~\cite{hadsell2006dimensionality} and triplet loss~\cite{schroff2015facenet} are two basic types of loss functions for deep metric learning. With a similar spirit of increasing and decreasing the distance between similar and dissimilar data samples, respectively, the former one takes pairs of sample as input while the latter is composed of triplets. Deep metric learning~\cite{fathi2017semantic} has proven effective in a wide variety of computer vision tasks, such as image retrieval~\cite{wang2014learning} and face recognition~\cite{schroff2015facenet}.

Although a few prior methods address the idea of metric learning in semantic segmentation, they only account for the local content from objects~\cite{harley2017segmentation} or instances~\cite{de2017semantic,bai2017deep,fathi2017semantic,kong2018grouppixels}. It$_{\!}$ is$_{\!}$ worth$_{\!}$ noting$_{\!}$ \cite{hwang2019segsort}$_{\!}$ also$_{\!}$ explores$_{\!}$ cross-image$_{\!}$ information$_{\!}$ of$_{\!}$ training data, \ie, leverage perceptual pixel groups for non-parametric pixel classification. Due to its clustering based metric learning strategy, \cite{hwang2019segsort} needs to retrieve extra labeled data for inference. Differently, our core idea, \ie, exploit inter-image pixel-to-pixel similarity to enforce global constraints on the embedding space, is conceptually novel and rarely explored before. It is executed by a compact training paradigm, which enjoys the complementary advantages of unary, pixel-wise cross-entropy loss and pair-wise, pixel-to-pixel contrast loss, without bringing any extra inference cost or modification to the base network during deployment.



\begin{figure*}[t]
	\vspace{-4pt}
	\begin{center}
		\includegraphics[width=\linewidth]{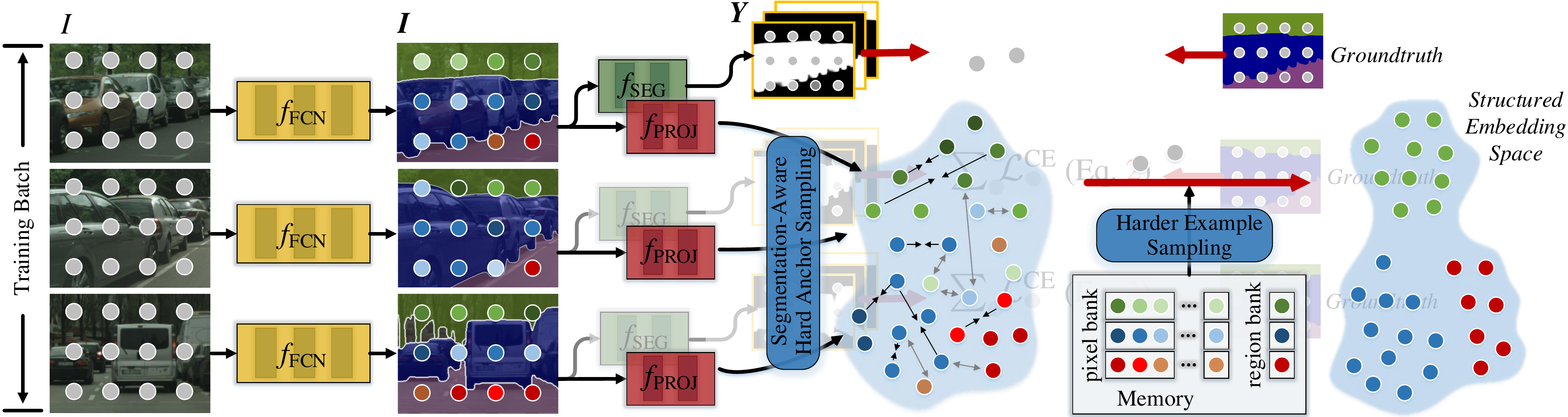}
		\put(-115,4){\scriptsize $\mathcal{M}$}
		\put(-150.5,84){\large$\sum\mathcal{L}^{\text{NCE}}$}
		\put(-110,84){(Eq.~\ref{eq:pNCE})}
		\put(-205,115.5){ \large$\sum~\!\mathcal{L}^{\text{CE}}$}
		\put(-164.5,115.5){(Eq.~\ref{eq:CE})}
	\end{center}
	\vspace{-16pt}
	\captionsetup{font=small} \caption{\small\textbf{Detailed illustration} of our pixel-wise contrastive learning based semantic segmentation network architecture.}
	\vspace{-10pt}
	\label{fig:framework}
\end{figure*}
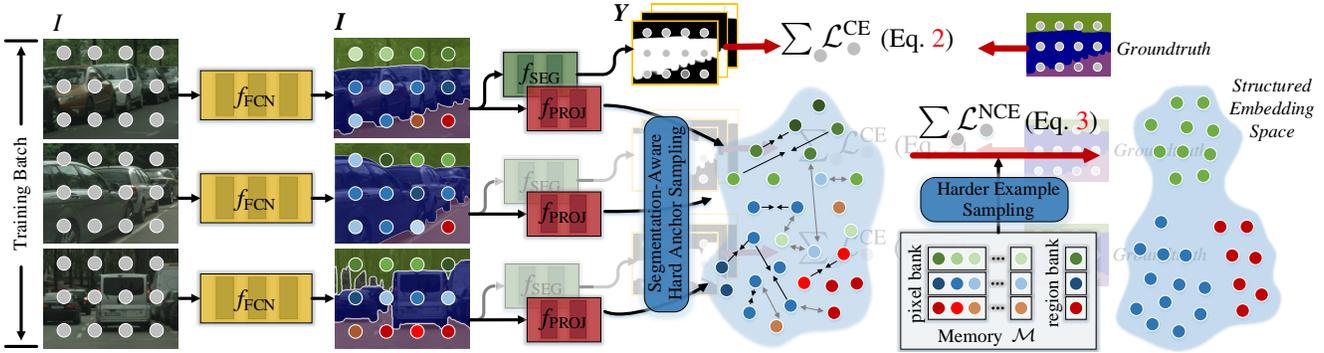

\vspace{-2pt}
\section{Methodology}
\vspace{-2pt}
Before detailing our supervised pixel-wise contrastive algorithm for semantic segmentation (\S\ref{sec:pc}), we first introduce the contrastive formulation in unsupervised visual representation learning and the notion of memory bank (\S\ref{sec:ic}).

\vspace{-2pt}
\subsection{Preliminaries}\label{sec:ic}
\vspace{-2pt}
\noindent\textbf{Unsupervised Contrastive Learning.} Unsupervised visual representation learning aims to learn a CNN encoder $f_{\text{CNN}}$ that  transforms each training image $I$ to a feature vector $\bm{v}\!=\!f_{\text{CNN}}(I)\!\in\mathbb{R}^D$, such that $\bm{v}$ best describes $I$. To achieve this goal, contrastive approaches conduct training by distinguishing a \textit{positive} (an augmented version of \textit{anchor} $I$) from several \textit{negatives} (images randomly drawn from the training set excluding $I$), based on the principle of similarity between samples. A popular loss function for contrastive learning, called InfoNCE~\cite{gutmann2010noise,oord2018representation}, takes the following form:\!\!\!
\vspace{-4pt}
\begin{equation}\small\label{eq:NCE}
\!\!\!\!\mathcal{L}^{\text{NCE}}_I\!=\!-\log\frac{\exp(\bm{v}\!\cdot\!\bm{v}^+/\tau)}{\exp(\bm{v}\!\cdot\!\bm{v}^+/\tau)
\!+\!\sum_{\bm{v}^-\in \mathcal{N}_I}\exp(\bm{v}\!\cdot\!\bm{v}^-/\tau)},\!\!
\vspace{-3pt}
\end{equation}
where $\bm{v}^+$ is an embedding of a positive for $I$, $\mathcal{N}_I$ contains embeddings of negatives, `$\cdot$' denotes the inner (dot) product, and $\tau\!>\!0$ is a temperature hyper-parameter. Note that all the embeddings in the loss function are $\ell_2$-normalized.


\noindent\textbf{Memory Bank.} As revealed by recent
studies~\cite{wu2018unsupervised,chen2020improved,he2020momentum}, a large set of negatives (\ie, $|\mathcal{N}_I|$) is critical in unsupervised contrastive representation learning.
As the number of negatives is  limited by the mini-batch size, recent contrastive methods  utilize large, external memories as a bank to store more navigate samples. Specifically, some methods~\cite{wu2018unsupervised} directly store the embeddings of all the training samples in the memory, however, easily suffering from asynchronous update. Some others choose to keep a queue of the last few batches~\cite{wang2020cross,chen2020improved,he2020momentum} as memory. In~\cite{chen2020improved,he2020momentum}, the stored embeddings are even updated on-the-fly through a momentum-updated version of the encoder network $f_{\text{CNN}}$.

%
%
%
%


\subsection{Supervised Contrastive Segmentation}\label{sec:pc}
\noindent\textbf{Pixel-Wise Cross-Entropy Loss.} In the context of semantic segmentation, each pixel $i$ of an image $I$ has to be classified into a semantic class $c\!\in\!\mathcal{C}$. Current approaches typically  cast this task as a pixel-wise classification problem. Specifically, let $f_{\text{FCN\!}}$ be an FCN encoder (\eg, ResNet~\cite{he2016deep}), that produces a dense feature $\bm{I}\!\in\!\mathbb{R}^{H\times W\times D\!}$ for $I$, from which the pixel embedding $\bm{i}\!\in\!\mathbb{R}^{D\!}$ of $i$ can be derived (\ie, $\bm{i}\!\in\!\bm{I}$). Then a segmentation head $f_{\text{SEG\!}}$ maps $\bm{I}$ into a categorical score map $\bm{Y}\!=\!f_{\text{SEG\!}}(\bm{I})\!\in\!\mathbb{R}^{H\times W\times |\mathcal{C}|\!}$.
Further let ${\bm{y}}\!=\![y_1, \cdots\!, y_{|\mathcal{C}|}]\!\in\!\mathbb{R}^{|\mathcal{C}|\!}$ be the \textit{unnormalized} score vector (termed as \textit{logit}) for pixel $i$, derived from $\bm{Y}$, \ie, ${\bm{y}}\!\in\!\bm{Y}$. Given ${\bm{y}}$ for pixel $i$ w.r.t its groundtruth label $\bar{c}\!\in\!\mathcal{C}$, the cross-entropy loss is optimized with \texttt{softmax} (\textit{cf.} Fig.~\ref{fig:framework}):\!\!
\vspace{-3pt}
\begin{equation}\small\label{eq:CE}
\mathcal{L}^{\text{CE}}_i=-\bm{1}^{\!\top}_{\bar{c}}\log (\texttt{softmax}({\bm{y}})),
\vspace{-3pt}
\end{equation}
where $\bm{1}_{\bar{c}}$ denotes the one-hot encoding of $\bar{c}$,  the logarithm is defined
as element-wise, and $\texttt{softmax}({y}_c)\!=\!\frac{\exp({y}_c)}{\sum_{c'\!=\!1}^{|\mathcal{C}|}\exp(y_{c'})}$.
Such training objective design mainly suffers from two limitations. \textbf{1)} It penalizes pixel-wise predictions independently but ignores relationship between pixels~\cite{2019_zhao_rmi}. \textbf{2)} Due to the use of \texttt{softmax}, the loss only depends on the relative relation among logits and cannot directly supervise on the learned representations~\cite{pang2020rethinking}. These two issues were rarely noticed; only a few structure-aware losses are designed to address \textbf{1)}, by considering pixel affinity~\cite{ke2018adaptive}, optimizing intersection-over-union measurement~\cite{berman2018lovasz}, or maximizing the mutual information between the groundtruth and prediction map~\cite{2019_zhao_rmi}. Nevertheless, these alternative losses only consider the dependencies between pixels within an image (\ie, local context),  regardless of the semantic correlations between pixels across images (\ie, global structure).


\noindent\textbf{Pixel-to-Pixel Contrast.} In this work, we develop a pixel-wise contrastive learning method that addresses both \textbf{1)} and \textbf{2)}, through regularizing the embedding space and exploring the global structures of training data. We first extend Eq.~(\ref{eq:NCE}) to our supervised, dense image prediction setting. Basically, the data samples in our contrastive loss computation are training image pixels. In addition, for a pixel $i$ with its groundtruth semantic label $\bar{c}$, the positive samples are other pixels also belonging to the class $\bar{c}$, while the negatives are the pixels belonging to the other classes $\mathcal{C}\backslash \bar{c}$. Our supervised, pixel-wise contrastive loss is defined as:
\vspace{-2pt}
\begin{equation}\small\label{eq:pNCE}
\!\!\!\!\mathcal{L}^{\text{NCE}}_i\!=\!\frac{1}{|\mathcal{P}_i|}\!\!\sum_{\bm{i}^+\in\mathcal{P}_i\!\!}\!\!\!-_{\!}\log\frac{\exp(\bm{i}\!\cdot\!\bm{i}^{+\!\!}/\tau)}{\exp(\bm{i}\!\cdot\!\bm{i}^{+\!\!}/\tau)
+\!\sum\nolimits_{\bm{i}^{-\!}\in\mathcal{N}_i\!}\!\exp(\bm{i}\!\cdot\!\bm{i}^{-\!\!}/\tau)},\!\!
\vspace{-2pt}
\end{equation}
where $\mathcal{P}_i$ and $\mathcal{N}_i$ denote pixel embedding collections of the positive and negative samples, respectively, for pixel $i$. Note that the positive/negative samples and the anchor $i$ are not restricted to being from a same image. As Eq.~(\ref{eq:pNCE}) shows, the purpose of such  \textit{pixel-to-pixel contrast} based loss design is to learn an embedding space, by pulling the same class pixel samples close and by pushing different class samples apart.

\begin{figure}[t]
	\begin{center}
		\includegraphics[width=\linewidth]{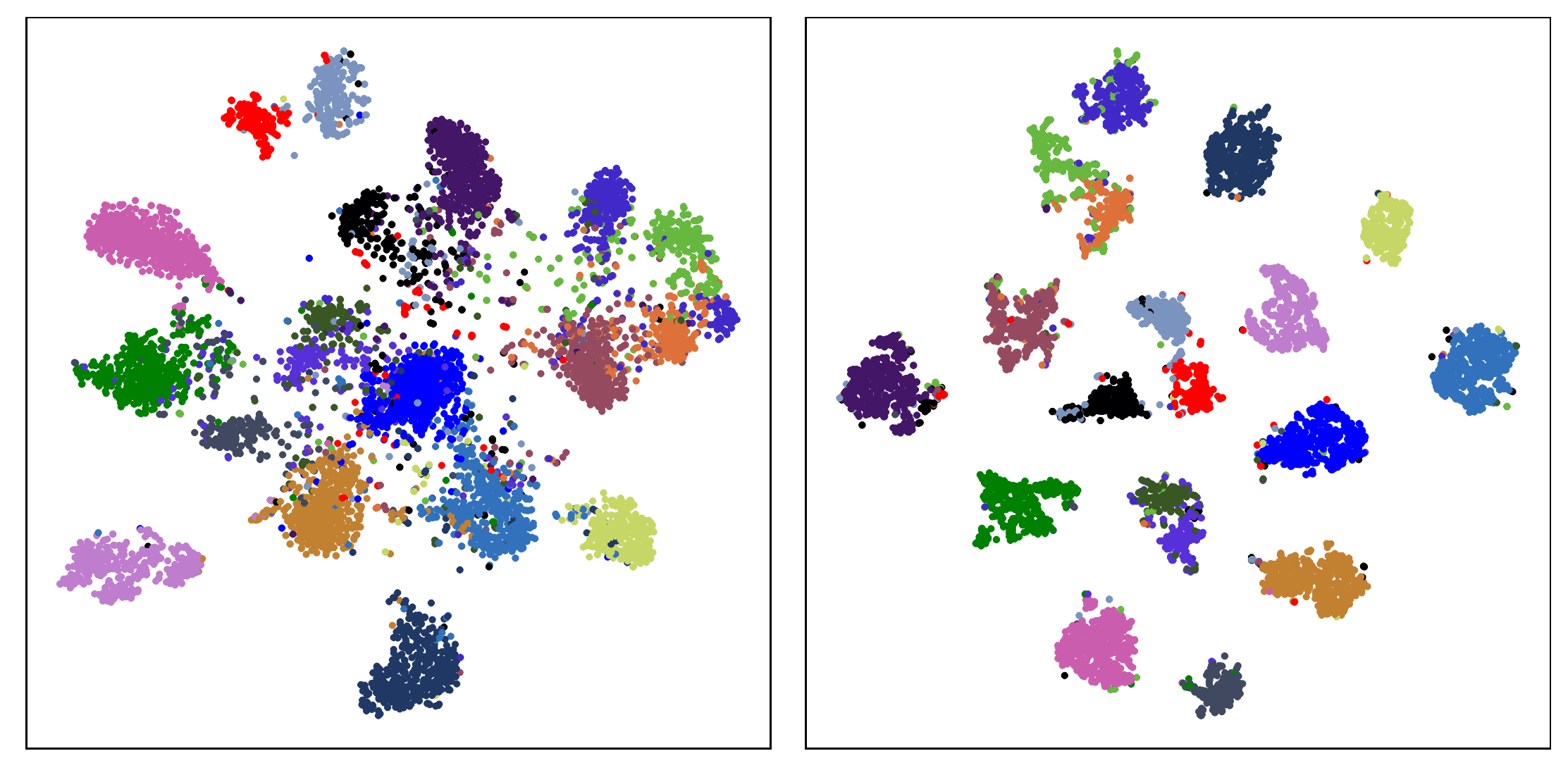}
	\end{center}
	\vspace{-18pt}
	\captionsetup{font=small}
	\caption{\small Visualization of features learned with (left) the pixel-wise entropy loss (\ie, $\mathcal{L}^{\text{CE}}$ in Eq.~\eqref{eq:CE}) and (right) our pixel contrast based optimization objective (\ie, $\mathcal{L}^{\text{SEG}}$ in Eq.~\eqref{eq:com}) on Cityscapes \texttt{val}~\cite{cordts2016cityscapes}. Features are colored according to class labels. As seen, the proposed $\mathcal{L}^{\text{SEG}}$ begets a well-structured semantic feature space.}
	\vspace{-10pt}
	\label{fig:tsne}
\end{figure}

The pixel-wise cross-entropy loss in Eq.~(\ref{eq:CE}) and our contrastive loss in Eq.~(\ref{eq:pNCE}) are complementary to each other; the former lets segmentation networks learn discriminative pixel features that are meaningful for classification, while the latter helps to regularize the embedding space with improved intra-class compactness and inter-class separability through \textit{explicitly} exploring global semantic relationships between pixel samples. Thus the overall training target is:
\vspace{-1pt}
\begin{equation}\small\label{eq:com}
\mathcal{L}^{\text{SEG}}=\sum\nolimits_i\big(\mathcal{L}^{\text{CE}}_i + \lambda\mathcal{L}^{\text{NCE}}_i\big),
\vspace{-1pt}
\end{equation}
where $\lambda\!>\!0$ is the coefficient. As shown in Fig.~\!\ref{fig:tsne}, the learned pixel embeddings by $\mathcal{L}^{\text{SEG}\!}$ become more compact and well separated.  This suggests that, by enjoying the advantage of unary cross-entropy loss and pair-wise metric loss, segmentation network can generate more discriminative features, hence producing more promising results. Quantitative analyses are later provided in \S\ref{sec:ma} and \S\ref{sec:qr}.

\noindent\textbf{Pixel-to-Region Contrast.} As stated in \S\ref{sec:ic}, memory is a critical technique that helps contrastive learning to make use of massive data to learn good representations. However, since there are vast numbers of pixel samples in our dense prediction setting and most of them are redundant (\ie, sampled from harmonious object regions), directly storing all the training pixel samples, like traditional memory~\cite{chen2020simple}, will greatly slow down the learning process. Maintaining several last batches in a queue, like~\cite{wang2020cross,chen2020improved,he2020momentum}, is also not a good choice, as recent batches only contain a limited number of images, reducing the diversity of pixel samples. Thus we choose to maintain a pixel queue per category. For each category, only a small number, \ie, $V$, of pixels are randomly selected from each image in the latest mini-batch, and pulled into the queue, with a size of $T\!\gg\!V$. In practice we find this strategy is very efficient and effective, but the under-sampled pixel embeddings are too sparse to fully capture image content.  Therefore, we further build a region memory bank that stores more representative embeddings absorbed from image segments (\ie, semantic regions).

Specifically, for a segmentation dataset with a total of $N$ training images and $|\mathcal{C}|$ semantic classes, our region memory is built with size $|\mathcal{C}|\!\times\!N\!\times\!D$, where $D$ is the dimension of pixel embeddings. The $(\bar{c}, n)$-th element in the region memory is a $D$-dimensional feature vector obtained by average pooling all the embeddings of pixels labeled as $\bar{c}$ category in the $n$-th image. The region memory brings two advantages: 1) store more representative ``pixel'' samples with low memory consumption; and 2) allow our pixel-wise contrastive loss (\textit{cf.}$_{\!}$ Eq.\!~(\ref{eq:pNCE})) to further explore pixel-to-region relations. With regard to 2), when computing Eq.\!~(\ref{eq:pNCE}) for an anchor pixel $i$ belonging to $\bar{c}$ category, stored region embeddings with the same class $\bar{c}$ are viewed as positives, while the region embeddings with other classes $\mathcal{C}\backslash \bar{c}$ are negatives. For the pixel memory, the size is $|\mathcal{C}|\!\times\!T\!\times\! D$. Therefore, for the whole memory (denoted as $\mathcal{M}$), the total size is $|\mathcal{C}|\!\times\! (N\!+\!T) \!\times\! D$. We examine the design of $\mathcal{M}$ in \S\ref{sec:ma}. In the following sections, we will not distinguish pixel and region embeddings in $\mathcal{M}$, unless otherwise specified.


\noindent\textbf{Hard Example Sampling.} Prior research~\cite{schroff2015facenet,kaya2019deep,khosla2020supervised,robinson2020contrastive,kalantidis2020hard} found that, in addition to loss designs and the amount of training samples,  the discriminating power of the training samples is crucial for metric learning. 
Considering our case, the gradient of the pixel-wise contrastive loss (\textit{cf.} Eq.~(\ref{eq:pNCE})) w.r.t.~the anchor embedding $\bm{i}$ can be given as:
\vspace{-2pt}
\begin{equation}\small\label{eq:gpNCE}
\!\!\frac{\partial\mathcal{L}^{\text{NCE}}_i}{\partial\bm{i}}\!=\!-_{\!}\frac{1}{\tau|\mathcal{P}_i|}\!\sum\nolimits_{\bm{i}^{+\!}\in\mathcal{P}_i\!\!\!}\Big((1-p_{i^+\!})\cdot\bm{i}{^+\!}-\sum\nolimits_{\bm{i}^{-\!}\in\mathcal{N}_i\!}\!p_{i^-}\cdot\bm{i}^{-\!}\Big),\!\!
\end{equation}
where $p_{i^{+\!/\!-\!\!}}\!\in\![0,1]$ denotes the matching probability between a positive/negative ${i}^{+\!/\!-\!}$ and the anchor ${i}$, \ie, $p_{i^{+\!/\!-\!\!}}\!=\!\!\frac{\exp(\bm{i}\cdot\bm{i}^{+\!/\!-\!\!}/\tau)}{\sum_{\bm{i}'^{\!}\in\mathcal{P}_{\!i}\cup\mathcal{N}_{\!i}\!}\exp(\bm{i}\cdot\bm{i}'^{\!}/\tau)}$. We$_{\!}$ view$_{\!}$ the$_{\!}$ negatives$_{\!}$ with$_{\!}$ dot$_{\!}$ products$_{\!}$  (\ie, $\bm{i}\!\cdot\!\bm{i}^-$) closer$_{\!}$  to$_{\!}$  $1$$_{\!}$  to$_{\!}$  be$_{\!}$  {\textit{harder}}, \ie, negatives which are similar to the anchor ${i}$.
Similarly, the positives with dot products (\ie, $\bm{i}\!\cdot\!\bm{i}^+$) closer to $-1$ are considered as
{\textit{harder}}, \ie, positives$_{\!}$  which$_{\!}$  are$_{\!}$  dissimilar$_{\!}$  to$_{\!}$  $i$.$_{\!}$  We~can~find that, harder negatives bring more gradient contributions, \ie, $p_{i^-}$, than easier negatives. This principle also holds true for positives, whose gradient contributions are $1_{\!}-_{\!}p_{i^+\!}$. Kalantidis \etal~\cite{kalantidis2020hard} further indicate that, as training progresses,$_{\!}$ more$_{\!}$ and$_{\!}$ more$_{\!}$ negatives$_{\!}$ become$_{\!}$ too simple to provide significant contributions to the unsupervised contrastive loss (\textit{cf.}$_{\!}$ Eq.\!~(\ref{eq:NCE})). This also happens in our supervised setting (\textit{cf.}$_{\!}$ Eq.\!~(\ref{eq:pNCE})), for both negatives and positives. To remedy this problem,~we propose the following sampling strategies:

\begin{itemize}[leftmargin=*]
	\setlength{\itemsep}{0pt}
	\setlength{\parsep}{-2pt}
	\setlength{\parskip}{-0pt}
	\setlength{\leftmargin}{-10pt}
	\vspace{-4pt}
	\item \textit{Hardest$_{\!}$ Example$_{\!}$ Sampling.}$_{\!}$ Inspired$_{\!}$ by$_{\!}$ hardest negative mining in metric learning~\cite{bucher2016hard}, we first design a ``hardest example sampling'' strategy: for each anchor pixel embedding $\bm{i}$, only sampling top-$K$ hardest negatives and positives from the memory bank $\mathcal{M}$, for the computation of the pixel-wise contrastive loss (\ie,  $\mathcal{L}^{\text{NCE}}$ in~Eq.\!~(\ref{eq:pNCE})).
	\item \textit{Semi-Hard$_{\!}$ Example$_{\!}$ Sampling.}$_{\!}$ Some$_{\!}$ studies$_{\!}$ propose$_{\!}$~to$_{\!}$ make use of harder negatives, as$_{\!}$ optimizing with$_{\!}$~the$_{\!}$~hardest negatives for metric learning likely leads to bad local minima~\cite{schroff2015facenet,xie2020delving,frankle2020all}. Thus we further design~a~``semi-hard example sampling'' strategy: for each anchor embedding $\bm{i}$, we first collect top $10\%$ nearest negatives (resp. top $10\%$ farthest positives) from the memory bank $\mathcal{M}$, from which we randomly then sample $K$ negatives (resp. $K$ positives) for our contrastive loss computation.
	\item \textit{Segmentation-Aware Hard Anchor Sampling.}  Rather than mining informative positive and negative examples, we develop an anchor sampling strategy. We treat the  categorization ability of an anchor embedding as its importance during contrastive learning. This leads to ``segmentation-aware hard anchor sampling'': the pixels with incorrect predictions, \ie, $c\!\neq\!\bar{c}$, are treated as \textit{hard anchors}. For the contrastive loss computation (\textit{cf}.~Eq.\!~(\ref{eq:pNCE})), half of the anchors are randomly sampled and half are the hard ones. This anchor sampling strategy enables our contrastive learning to focus more on the pixels hard for classification, delivering more segmentation-aware embeddings.
	\vspace{-4pt}
\end{itemize}

In practice, we find ``semi-hard example sampling'' strategy performs better than ``hardest example sampling''. In addition, after employing ``segmentation-aware hard anchor sampling'' strategy, the segmentation performance can be further improved. See \S \ref{sec:ma} for related experiments.

\subsection{Detailed Network Architecture}\label{sec:nd}

Our algorithm has five major components (\textit{cf.}~Fig.\!~\ref{fig:framework}):
\begin{itemize}[leftmargin=*]
	\setlength{\itemsep}{0pt}
	\setlength{\parsep}{-2pt}
	\setlength{\parskip}{-0pt}
	\setlength{\leftmargin}{-10pt}
	\vspace{-6pt}
	\item \textit{FCN Encoder}, $f_{\text{FCN}}$, which maps each input image $I$ into dense embeddings $\bm{I}\!=\!f_{\text{FCN}}(I)\!\in\!\mathbb{R}^{H\!\times\!W\!\times\!D}$. In our algorithm, any FCN backbones can be used to implement $f_{\text{FCN}}$ and we test two commonly used ones, \ie, ResNet~\cite{he2016deep} and HRNet~\cite{wang2020deep}, in our experiments.
	
\item \textit{Segmentation Head}, $f_{\text{SEG}}$, that projects $\bm{I}$ into a score map $\bm{Y}\!\!=\!\!f_{\text{SEG}}(\bm{I})\!\in\!\mathbb{R}^{H\times\!W\!\times\!|\mathcal{C}|}$. We conduct evaluations using different segmentation heads in mainstream methods (\ie, DeepLabV3~\cite{chen2017rethinking}, HRNet~\cite{wang2020deep}, and OCR~\cite{yuan2020object}).

	\item \textit{Project Head}, $f_{\text{PROJ}}$, which maps each high-dimensional  pixel embedding $\bm{i}\!\in\!\bm{I}$ into a 256-\textit{d} $\ell_2$-normalized feature vector\!~\cite{chen2020simple}, for the computation of the contrastive loss $\mathcal{L}^{\text{NCE}}$. $f_{\text{PROJ}}$ is implemented as two $1\!\times1\!$  convolutional layers with \texttt{ReLU}. Note that the project head is only applied during training and is removed at inference time. Thus it does not introduce any changes to the segmentation network or extra computational cost in deployment.
	\item \textit{Memory Bank}, $\mathcal{M}$, which consists of two parts that store pixel and region embeddings, respectively. For each training image, we sample $V\!=\!10$ pixels per class. For each class, we set the size of the pixel queue as $T\!=\!10N$. 
The memory bank is also discarded after training.
	\item \textit{Joint Loss}, $\mathcal{L}^{\text{SEG}\!}$ (\textit{cf.} Eq.~\eqref{eq:com}), that takes the power of representation learning (\ie, $\mathcal{L}^{\text{CE}}$ in Eq.~\eqref{eq:CE}) and metric learning (\ie, $\mathcal{L}^{\text{NCE}}$ in Eq.~\eqref{eq:pNCE}) for more distinct segmentation feature learning. In practice, we find our method is not sensitive to the coefficient $\lambda$ (\eg, when $\lambda\!\in\![0.1,1]$) and empirically set $\lambda$ as 1.  For $\mathcal{L}^{\text{NCE}}$ in Eq.~\eqref{eq:pNCE}, we set the temperature $\tau$ as $0.1$. For sampling, we find  ``semi-hard example sampling'' + ``segmentation-aware hard anchor sampling'' performs the best and set the numbers of sampled instances (\ie, $K$) as $1,\!024$ and $2,\!048$ for positive and negative, respectively. For each mini-batch, 50 anchors are sampled per category (half are randomly sampled and the other half are segmentation-hard ones).
	\vspace{-4pt}
\end{itemize}


\vspace{-2pt}
\section{Experiment}\label{sec:ex}
\vspace{-2pt}
\subsection{Experimental Setup}
\vspace{-2pt}
\noindent\textbf{Datasets.}~Our experiments are conducted on four datasets:\!\!
\begin{itemize}[leftmargin=*]
	\setlength{\itemsep}{0pt}
	\setlength{\parsep}{-2pt}
	\setlength{\parskip}{-0pt}
	\setlength{\leftmargin}{-10pt}
	\vspace{-6pt}
	\item \textbf{Cityscapes}~\cite{cordts2016cityscapes} has $5,\!000$  finely annotated urban scene images, with $2,\!975/500/1,\!524$ for \texttt{train}/\texttt{val}/\texttt{test}. The segmentation performance is reported on $19$  challenging categories, such as person, sky, car, and building.
	
	\item \textbf{PASCAL-Context}~\cite{mottaghi2014role} contains $4,\!998$ and $5,\!105$ images in \texttt{train} and \texttt{test} splits, respectively, with precise annotations of $59$ semantic categories.

	\item \textbf{COCO-Stuff}~\cite{caesar2018coco} consists of $10,\!000$ images gathered from COCO~\cite{lin2014microsoft}. It is split into $9,\!000$ and $1,\!000$ images for \texttt{train} and \texttt{test}. It provides rich annotations for 80 object classes and 91 stuff classes.	
	
	\item \textbf{CamVid}~\cite{brostow2009semantic} has $367/101/233$ images for \texttt{train}/\texttt{val}/ \texttt{test}, with $11$ semantic labels in total. 
	\vspace{-4pt}
\end{itemize}

\noindent\textbf{Training.} As mentioned in \S\ref{sec:nd}, various backbones (\ie, ResNet\!~\cite{he2016deep} and HRNet\!~\cite{wang2020deep}) and segmentation networks (\ie, DeepLabV3\!~\cite{chen2017rethinking}, HRNet\!~\cite{wang2020deep}, and OCR\!~\cite{yuan2020object})  are exploited in our experiments to thoroughly validate the proposed algorithm. We follow conventions~\cite{wang2020deep,yuan2020object,choi2020cars,yin2020disentangled} for training hyper-parameters. For fairness, we initialize all backbones using corresponding weights pretrained on ImageNet~\cite{ImageNet}, with the remaining layers being randomly initialized. For data augmentation, we use color jitter, horizontal flipping and random scaling with a factor in $[0.5,~2]$. We use SGD as our optimizer, with a momentum $0.9$ and weight decay $0.0005$. We adopt the polynomial annealing policy~\cite{chen2017rethinking} to schedule the learning rate, which is multiplied by $(1\!-\!\frac{iter}{total\_iter})^{power\!}$ with $power\!=\!0.9$. Moreover, for Cityscapes, we use a mini-batch size of $8$, and an initial learning rate of $0.01$. All the training images are augmented by random cropping from $1024\!\times\!2048$ to $512\!\times\!1024$.
For the experiments on \texttt{test}, we follow~\cite{wang2020deep} to train the model for $100${K} iterations. Note that we do not use any extra training data (\eg, Cityscapes coarse~\cite{cordts2016cityscapes}). For PASCAL-Context and COCO-Stuff, we opt a mini-batch size of $16$, an initial learning rate of $0.001$, and crop size of $520\!\times\!520$. We train for $60${K} iterations over their $\texttt{train}$ sets. For CamVid, we train the model for $6${K} iterations, with batch size $16$, learning rate $0.02$ and original image size.


\noindent\textbf{Testing.} Following general protocol~\cite{wang2020deep,yuan2020object,shen2020ranet}, we average the segmentation results over  multiple scales with flipping, \ie, the scaling factor is $0.75$ to $2.0$ (with intervals of $0.25$) times of the original image size. Note that, during testing, there is no any change or extra inference step introduced to the base segmentation models, \ie, the projection head, $f_{\text{PROJ}}$,  and memory bank, $\mathcal{M}$, are directly discarded.

\noindent\textbf{Evaluation Metric.}  Following the standard setting, mean intersection-over-union (mIoU) is used for evaluation.

\noindent\textbf{Reproducibility.} Our model is implemented in PyTorch and trained on four NVIDIA Tesla V100 GPUs with a 32GB memory per-card. Testing is conducted on the same machine. Our implementations will be publicly released.


\vspace{-2pt}
\subsection{Diagnostic Experiment}\label{sec:ma}
We first study the efficacy of our core ideas and essential model designs, over Cityscapes \texttt{val}~\cite{cordts2016cityscapes}. We adopt HRNet~\cite{wang2020deep} as our base segmentation network (denoted as ``Baseline (\textit{w/o} contrast)'' in Tables \ref{table:contrastive}-\ref{table:sampling}). To perform extensive ablation experiments, we train each model for $40$K iterations while keeping other hyper-parameters unchanged.

\begin{table}
	\centering
	\small
	\resizebox{0.48\textwidth}{!}{
		\setlength\tabcolsep{8pt}
		\renewcommand\arraystretch{1.0}
		\begin{tabular}{c||c|c}
			\hline\thickhline
			\rowcolor{mygray}
			Pixel Contrast  & Backbone & mIoU (\%)  \\ \hline\hline
			Baseline (\textit{w/o} contrast) & HRNetV2-W48   & 78.1\\\hline
			Intra-Image Contrast & HRNetV2-W48   & 78.9  \color{mygreen}{({+0.8})}\\
			Inter-Image Contrast & HRNetV2-W48  & \textbf{81.0} \color{mygreen}{(\textbf{+2.9})} \\ \hline
		\end{tabular}
	}
	\captionsetup{font=small}
	\caption{\small \textbf{Comparison of different contrastive mechanisms}  on Cityscapes \texttt{val}~\cite{cordts2016cityscapes}. See \S\ref{sec:ma} for more details.}
	\label{table:contrastive}
	\vspace{-8pt}
\end{table}

\begin{table}
	\centering
	\small
	\resizebox{0.48\textwidth}{!}{
		\setlength\tabcolsep{8pt}
		\renewcommand\arraystretch{1.0}
		\begin{tabular}{c||c|c}
			\hline\thickhline
			\rowcolor{mygray}
			Memory & Backbone & mIoU (\%)  \\ \hline\hline
			Baseline (\textit{w/o} contrast) & HRNetV2-W48   & 78.1\\\hline
			Mini-Batch (\textit{w/o} memory) & HRNetV2-W48 & 79.8 \color{mygreen}{({+1.7})} \\
			Pixel Memory & HRNetV2-W48 & 80.5 \color{mygreen}{({+2.6})} \\
			Region Memory & HRNetV2-W48 & 80.2 \color{mygreen}{({+2.1})} \\ \hline
			Pixel + Region Memory & HRNetV2-W48 & \textbf{81.0}  \color{mygreen}{(\textbf{+2.9})} \\ \hline
		\end{tabular}
	}
	\captionsetup{font=small}
	\caption{\small \textbf{Comparison of different memory bank designs}  on Cityscapes \texttt{val}~\cite{cordts2016cityscapes}. See \S\ref{sec:ma} for more details.}
	\label{table:memory}
	\vspace{-12pt}
\end{table}

\noindent\textbf{Inter-Image \textit{vs.}~\!Intra-Image Pixel Contrast.}
We first investigate the effectiveness of our core idea of inter-image pixel contrast. As shown in Table~\ref{table:contrastive}, additionally considering cross-image pixel semantic relations (\ie, ``Inter-Image Contrast'') in segmentation network learning leads to a substantial performance gain (\ie, $\textbf{2.9}\%$), compared with ``Baseline (\textit{w/o} contrast)''. In addition, we develop another baseline, ``Intra-Image Contrast'', which only samples pixels from same images during the contrastive loss (\ie, $\mathcal{L}^{\text{NCE}}$ in Eq.~(\ref{eq:gpNCE})) computation. The results in Table~\ref{table:contrastive} suggest that, although ``Intra-Image Contrast'' also boosts the performance over ``Baseline (\textit{w/o} contrast)'' (\ie, 78.1\%$\rightarrow$78.9\%), ``Inter-Image Contrast'' is more favored.

\begin{table}
	\centering
	\small
	\resizebox{0.48\textwidth}{!}{
		\setlength\tabcolsep{8pt}
		\renewcommand\arraystretch{1.0}
		\begin{tabular}{c|c||c|c}
			\hline\thickhline
			\rowcolor{mygray}
			\multicolumn{2}{c||}{Sampling}  & & \\
            \cline{1-2}
            \rowcolor{mygray}
            Anchor &Pos./Neg. &\multirow{-2}*{Backbone} &\multirow{-2}*{mIoU (\%)}  \\ \hline\hline
            \multicolumn{2}{c||}{Baseline (\textit{w/o} contrast)} & HRNetV2-W48   & 78.1\\\hline
			   & Random &HRNetV2-W48  & 79.3 \color{mygreen}{({+1.2})}  \\
Random  & Hardest &HRNetV2-W48  & 79.4 \color{mygreen}{({+1.3})}\\
			   & Semi-Hard &HRNetV2-W48  & 80.1 \color{mygreen}{({+2.0})} \\ \hline
\multirow{3}{*}{\tabincell{c}{Seg.-aware \\hard}}& Random & HRNetV2-W48  & 80.2 \color{mygreen}{({+2.1})} \\
  				& Hardest &HRNetV2-W48  & 80.5 \color{mygreen}{({+2.4})} \\
 				& Semi-Hard &HRNetV2-W48  & \textbf{81.0} \color{mygreen}{(\textbf{+2.9})}\\\hline
		\end{tabular}
	}
	\captionsetup{font=small}
	\caption{\small \textbf{Comparison of different hard example sampling strategies} on Cityscapes \texttt{val}~\cite{cordts2016cityscapes}.  See \S\ref{sec:ma} for more details.}
	\label{table:sampling}
	\vspace{-8pt}
\end{table}

\noindent\textbf{Memory Bank.} We next validate the design of our memory bank. The results are summarized in Table~\ref{table:memory}. Based on ``Baseline (\textit{w/o} contrast)'', we first derive a variant, ``Mini-Batch \textit{w/o memory}'': only compute pixel contrast within each mini-batch, without outside memory. It gets $79.8\%$ mIoU. We then provision this variant with pixel and region memories separately, and observe consistent performance gains ($79.8\%\!\rightarrow\!80.5\%$ for pixel memory and $79.8\%\!\rightarrow\!80.2\%$ for region memory). This evidences that \textbf{i)} leveraging more pixel samples during contrastive learning leads to better pixel embeddings; and \textbf{ii)} both pixel-to-pixel and pixel-to-region relations are informative cues. Finally, after using both the two memories, a higher score (\ie, $81.0\%$) is achieved, revealing \textbf{i)} the effectiveness of our memory design; and \textbf{ii)} necessity of comprehensively considering both pixel-to-pixel contrast and pixel-to-region contrast.

\noindent\textbf{Hard Example Mining.}
Table~\ref{table:sampling} presents a comprehensive examination of various hard example mining strategies proposed in \S\ref{sec:pc}. Our main
observations are the following: \textbf{i)} For positive/negative sampling, mining meaningful pixels (\ie, ``hardest'' or ``semi-hard'' sampling), rather than ``random'' sampling, is indeed useful; \textbf{ii)} Hence, ``semi-hard'' sampling is more favored, as it improves the robustness of training by avoiding overfitting outliers in the training set. This corroborates related observations in unsupervised setting~\cite{wu2020mutual} and indicates that segmentation may benefit from more intelligent sample treatment; and \textbf{iii)} For anchor sampling, ``seg.-aware hard'' strategy further improves the performance (\ie, 80.1\%$\rightarrow$81.0\%) over ``random'' sampling only. This suggests that exploiting task-related signals in supervised metric learning may help develop better segmentation solutions, which has remained relatively
untapped.


\begin{figure*}[t]
	\begin{center}
		\includegraphics[width=\linewidth]{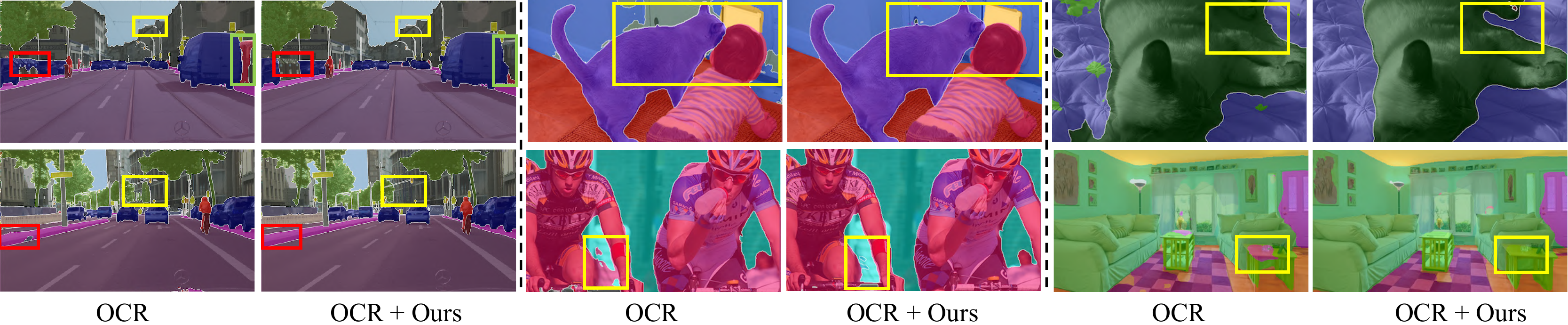}
	\end{center}
	\vspace{-16pt}
	\captionsetup{font=small}
	\caption{\small\textbf{Visual comparisons} between \textit{OCR}~\cite{yuan2020object} and \textit{OCR+Ours} (from left to right: Cityscapes, PASCAL-Context, COCO-Stuff).}
	\vspace{-10pt}
	\label{fig:visual}
\end{figure*}

\begin{table}
	\centering
	\small
	\resizebox{0.47\textwidth}{!}{
		\setlength\tabcolsep{12pt}
		\renewcommand\arraystretch{1.0}
		\begin{tabular}{r||c|c}\thickhline
			\rowcolor{mygray}
			Model & Backbone & mIoU (\%)  \\ \hline\hline
			\multicolumn{3}{l}{\textit{Model learned on Cityscapes \texttt{train}}} \\  \hline
			PSPNet$_{17}$~\cite{zhao2017pyramid} & D-ResNet-101 & 78.4 \\
			PSANet$_{18}$~\cite{zhao2018psanet} & D-ResNet-101 & 78.6 \\
			PAN$_{18}$~\cite{li2018pyramid} & D-ResNet-101 & 78.6  \\
			AAF$_{18}$~\cite{ke2018adaptive} & D-ResNet-101 & 79.1 \\\hline
			DeepLabV3$_{17}$~\cite{chen2017rethinking} & D-ResNet-101 & 78.1 \\
			DeepLabV3 $+~\textbf{\texttt{Ours}}$ & D-ResNet-101 & \textbf{79.2}  \color{mygreen}{(\textbf{+1.1}}) \\ \hline
			HRNetV2$_{20}$~\cite{wang2020deep} & HRNetV2-W48 & 80.4  \\
			HRNetV2$+~\textbf{\texttt{Ours}}$ & HRNetV2-W48 & \textbf{81.4} \color{mygreen}{(\textbf{+1.0}})   \\

			\hline\hline
			\multicolumn{3}{l}{\textit{Model learned on Cityscapes $\texttt{train+val}$}} \\  \hline
			DFN$_{18}$~\cite{yu2018learning} & D-ResNet-101 & 79.3 \\
			PSANet$_{18}$~\cite{zhao2018psanet} & D-ResNet-101 & 80.1 \\
			SVCNet$_{19}$~\cite{ding2019semantic} & D-ResNet-101 & 81.0 \\
			CPN$_{20}$~\cite{yu2020context} & D-ResNet-101 & 81.3 \\
			DANet$_{19}$~\cite{fu2019dual} & D-ResNet-101 & 81.5  \\
			
			ACF$_{19}$~\cite{zhang2019acfnet} & D-ResNet-101 & 81.8 \\
			DGCNet$_{19}$~\cite{zhang2019dual} & D-ResNet-101 & 82.0 \\
			HANet$_{20}$~\cite{choi2020cars} & D-ResNet-101 & 82.1 \\
			ACNet$_{19}$~\cite{fu2019adaptive} &  D-ResNet-101 & 82.3 \\
			\hline

			
			DeepLabV3$_{17}$~\cite{chen2017rethinking} & D-ResNet-101 & 79.4\\
			DeepLabV3 $+~\textbf{\texttt{Ours}}$ & D-ResNet-101 & \textbf{80.3} \color{mygreen}{(\textbf{+0.9})} \\ \hline
			HRNetV2$_{20}$~\cite{wang2020deep} & HRNetV2-W48 & 81.6  \\
			HRNetV2$+~\textbf{\texttt{Ours}}$&HRNetV2-W48 &  \textbf{82.5} \color{mygreen}{(\textbf{+0.9})} \\ \hline
			OCR$_{20}$~\cite{yuan2020object} & HRNetV2-W48 & 82.4   \\
			OCR$+~\textbf{\texttt{Ours}}$&HRNetV2-W48 &  \textbf{83.2}  \color{mygreen}{(\textbf{+0.8})} \\ \hline
		\end{tabular}
	}
	\captionsetup{font=small}
	\caption{\small \textbf{Quantitative  segmentation results} on Cityscapes \texttt{test}~\cite{cordts2016cityscapes}. D-ResNet-101 = Dilated ResNet-101.  See \S\ref{sec:qr}.}
	\label{table:city}
	\vspace{-8pt}
\end{table}

\subsection{Comparison to State-of-the-Arts}\label{sec:qr}

\noindent\textbf{Cityscapes\!~\cite{cordts2016cityscapes}.}
Table~\ref{table:city} lists the scores on Cityscapes \texttt{test}, under two widely used training settings~\cite{wang2020deep} (trained over \texttt{train} or \texttt{train+val}). Our method brings impressive gains over 3 strong baselines (\ie, DeepLabV3, HRNetV2,  and OCR), and sets a new state-of-the-art.

\begin{table}
	\centering
	\small
	\resizebox{0.44\textwidth}{!}{
		\setlength\tabcolsep{10pt}
		\renewcommand\arraystretch{1.0}
		\begin{tabular}{r||c|c}\thickhline
			\rowcolor{mygray}
			Model & Backbone & mIoU (\%) \\ \hline\hline
			DANet$_{19}$~\cite{fu2019dual} & D-ResNet-101 & 52.6  \\
			SVCNet$_{19}$~\cite{ding2019semantic} & D-ResNet-101 & 53.2 \\
			CPN$_{20}$~\cite{yu2020context} & D-ResNet-101 & 53.9 \\
			ACNet$_{19}$~\cite{fu2019adaptive} & D-ResNet-101 & 54.1 \\
			DMNet$_{19}$~\cite{he2019dynamic}&D-ResNet-101 & 54.4\\
			RANet$_{20}$~\cite{shen2020ranet} & ResNet-101 & 54.9  \\
			DNL$_{20}$~\cite{yin2020disentangled} & HRNetV2-W48 & 55.3 \\
			
			\hline\hline
			
			
			
			
			HRNetV2$_{20}$~\cite{wang2020deep} & HRNetV2-W48 & 54.0 \\
			HRNetV2$+~\textbf{\texttt{Ours}}$&HRNetV2-W48 &  \textbf{55.1} \color{mygreen}{\textbf{(+1.1)}} \\ \hline
			OCR$_{20}$~\cite{yuan2020object} & HRNetV2-W48 & 56.2  \\
			OCR$+~\textbf{\texttt{Ours}}$&HRNetV2-W48 &  \textbf{57.2} \color{mygreen}{\textbf{(+1.0)}}   \\ \hline
		\end{tabular}
	}
	\captionsetup{font=small}
	\caption{\small \textbf{Quantitative  segmentation results} on PASCAL-Context \texttt{test}~\cite{mottaghi2014role}.  D-ResNet-101 = Dilated ResNet-101. See \S\ref{sec:qr}.}
	\label{table:pascal}
	\vspace{-10pt}
\end{table}

\begin{table}
	\centering
	\small
	\resizebox{0.44\textwidth}{!}{
		\setlength\tabcolsep{12pt}
		\renewcommand\arraystretch{1.0}
		\begin{tabular}{r||c|c}\thickhline
			\rowcolor{mygray}
			Model & Backbone & mIoU (\%)  \\ \hline\hline
			SVCNet$_{19}$~\cite{ding2019semantic} & D-ResNet-101 & 39.6 \\
			DANet$_{19}$~\cite{fu2019dual} & D-ResNet-101 & 39.7  \\
			SpyGR$_{20}$~\cite{li2020spatial} & ResNet-101 & 39.9 \\
			ACNet$_{19}$~\cite{fu2019adaptive} & ResNet-101 & 40.1 \\
			\hline
			HRNetV2$_{20}$~\cite{wang2020deep} & HRNetV2-W48 & 38.7 \\
			HRNetV2$+~\textbf{\texttt{Ours}}$&HRNetV2-W48 &  \textbf{39.3} \color{mygreen}{\textbf{(+0.6)}} \\ \hline
			OCR$_{20}$~\cite{yuan2020object} & HRNetV2-W48 & 40.5   \\
			OCR$+~\textbf{\texttt{Ours}}$&HRNetV2-W48 &  \textbf{41.0} \color{mygreen}{\textbf{(+0.5)}}  \\ \hline
		\end{tabular}
	}
	\captionsetup{font=small}
	\caption{\small \textbf{Quantitative segmentation results} on COCO-Stuff \texttt{test}~\cite{caesar2018coco}. D-ResNet-101 = Dilated ResNet-101. See \S\ref{sec:qr}.}
	\label{table:coco}
	\vspace{-8pt}
\end{table}

\begin{table}
	\centering
	\small
	\resizebox{0.44\textwidth}{!}{
		\setlength\tabcolsep{12pt}
		\renewcommand\arraystretch{1.0}
		\begin{tabular}{r||c|c}\thickhline
			\rowcolor{mygray}
			Model & Backbone & mIoU (\%)  \\ \hline\hline
			DFANet$_{19}$~\cite{li2019dfanet} & Xception & 64.7 \\
			BiSeNet$_{18}$~\cite{yu2018bisenet} & D-ResNet-101 & 68.7 \\
			PSPNet$_{17}$~\cite{zhao2017pyramid} & D-ResNet-101 & 69.1 \\ \hline

			HRNetV2$_{20}$~\cite{wang2020deep} & HRNetV2-W48 & 78.5 \\
			HRNetV2$+~\textbf{\texttt{Ours}}$&HRNetV2-W48 &  \textbf{79.0} \color{mygreen}{\textbf{(+0.5)}}\\ \hline
			
			OCR$_{20}$~\cite{yuan2020object} & HRNetV2-W48 & 80.1 \\
			OCR$+~\textbf{\texttt{Ours}}$ & HRNetV2-W48 &  \textbf{80.5} \color{mygreen}{\textbf{(+0.4)}} \\ \hline
		\end{tabular}
	}
	\captionsetup{font=small}
	\caption{\small\textbf{Quantitative segmentation results} on CamVid \texttt{test}~\cite{brostow2009semantic}. D-ResNet-101$=$Dilated ResNet-101. See \S\ref{sec:qr}.}
	\label{table:camvid}
	\vspace{-14pt}
\end{table}

\noindent\textbf{PASCAL-Context~\cite{mottaghi2014role}.}
Table~\ref{table:pascal} presents comparison results on PASCAL-Context \texttt{test}. Our approach improves the performance of base networks by solid margins (\ie, $54.0\!\!\rightarrow\!\!55.1$ for HRNetV2, $56.2\!\!\rightarrow\!\!57.2$ for OCR). This is particularly impressive considering the fact that improvement on this extensively-benchmarked dataset is very hard.

\noindent\textbf{COCO-Stuff~\cite{caesar2018coco}.}
Table~\ref{table:coco} reports performance comparison of our method against seven competitors on COCO-Stuff \texttt{test}. As we find that \textit{OCR+\texttt{\textbf{Ours}}} yields a mIoU of $41.0\%$, which leads to a promising gain of $\textbf{0.5\%}$ over its counterpart (\ie, OCR with a $40.5\%$ mIoU). Besides, \textit{HRNetV2+\texttt{\textbf{Ours}}} outperforms HRNetV2 by $\textbf{0.6\%}$.

\noindent\textbf{CamVid\!~\cite{brostow2009semantic}.} Table~\ref{table:camvid} shows that our method also leads to improvements over HRNetV2 and OCR on CamVid \texttt{test}.

\noindent\textbf{Qualitative Results.}
\figref{fig:visual} depicts qualitative comparisons of \textit{OCR+\texttt{\textbf{Ours}}} against \textit{OCR} over representative examples from three datasets (\ie, Cityscapes, PASCAL-Context and COCO-Stuff). As seen, our method is capable of producing more accurate segments across various challenge scenarios. 



\section{Conclusion and Discussion}
In$_{\!}$ this$_{\!}$ paper, we propose a$_{\!}$ new$_{\!}$ supervised$_{\!}$ learning paradigm for semantic segmentation, enjoying the complementary advantages of unary classification and structured metric learning. Through pixel-wise contrastive learning, it investigates global semantic relations between training pixels, guiding pixel embeddings towards cross-image category-discriminative representations that eventually improve the segmentation performance. Our method generates promising results and shows great potential in a variety of dense image prediction tasks, such as pose estimation and medical image segmentation. It also comes with new challenges, in particular
regarding smart data sampling, metric learning loss design, class rebalancing during training, and multi-layer feature contrast. Given the massive number of technique breakthroughs over the past few years, we expect a flurry of innovation towards these promising directions.

%
%
%
%





\appendix

\setcounter{table}{0}
\renewcommand{\thetable}{A\arabic{table}}
\setcounter{figure}{0}
\renewcommand{\thefigure}{A\arabic{figure}}


\begin{table*}[t]
	\centering
	\small
	\resizebox{\textwidth}{!}{
		\setlength\tabcolsep{18pt}
		\renewcommand\arraystretch{1.0}
		\begin{tabular}{r||cc|c|c}
			\hline\thickhline
			\rowcolor{mygray}
			Loss  & Type & Context & Backbone & mIoU (\%)  \\ \hline\hline
			Cross-Entropy Loss & unary & local & HRNetV2-W48   & 78.1 \\
			+AAF Loss~\cite{ke2018adaptive} & pairwise & local &HRNetV2-W48 & 78.7 \\
			+RMI Loss~\cite{2019_zhao_rmi} & higher-order & local &HRNetV2-W48 & 79.8 \\
			+Lov{\'a}sz Loss~\cite{berman2018lovasz} & higher-order & local &HRNetV2-W48 & 80.3\\ \hline\hline
			+Contrastive Loss (\textbf{Ours})& pairwise & global &HRNetV2-W48  & {81.0} \\\hline\hline
			+AAF~\cite{ke2018adaptive} + Contrastive& - & - &HRNetV2-W48  & 81.0  \\
			+RMI~\cite{2019_zhao_rmi} + Contrastive& - & -&HRNetV2-W48  &  81.3 \\
			+Lov{\'a}sz~\cite{berman2018lovasz} + Contrastive & -& -&HRNetV2-W48  & 81.5 \\ \hline
		\end{tabular}
	}
	\captionsetup{font=small}
	\caption{\small \textbf{Comparison of different loss designs}  on Cityscapes \texttt{val}~\cite{cordts2016cityscapes}. See \S\ref{sec:loss} for more details.}
	\label{table:loss}
	\vspace{-6pt}
\end{table*}

\section{Comparison to Other Losses}\label{sec:loss}

We further study the effectiveness of our contrastive loss against representative semantic segmentation losses, including Cross-Entropy (CE) Loss, AAF loss~\cite{ke2018adaptive}, Lov{\'a}sz Loss~\cite{berman2018lovasz}, and RMI Loss~\cite{2019_zhao_rmi}.


For fair comparison, we examine each loss using HRNetV2~\cite{wang2020deep} as the base segmentation network, and train the loss jointly with CE on Cityscapes \texttt{train} for $40,\!000$ iterations with a mini-batch size of $8$. The results are reported in Table~\ref{table:loss}. We observe that all structure-aware losses outperform the standard CE loss. Notably, our contrastive loss achieves the best performance, outperforming the second-best Lov{\'a}sz loss by $\textbf{0.7\%}$, and the pairwise losses, \ie, RMI and AAF, by $\textbf{1.2\%}$ and $\textbf{2.3\%}$, respectively.

Additionally, Table~\ref{table:loss} reports results of each loss in combination with our contrastive loss. From a perspective of metric learning,  the CE loss can be viewed as a pixel-wise \textit{unary} loss that penalizes each pixel independently and ignores dependencies between pixels, while AAF is a \textit{pairwise} loss, which models the pairwise relations between spatially adjacent pixels. Moreover, the RMI and Lov{\'a}sz losses are \textit{higher-order} losses: the former one accounts for region-level mutual information, and the latter one  directly optimizes the intersection-over-union score over the pixel clique level. However, all these existing loss designs are defined within individual images, capturing \textit{local} context/pixel relations only. Our contrastive loss, as it explores \textit{pairwise} pixel-to-pixel dependencies, is also a \textit{pairwise} loss. But it is computed over the whole training dataset, addressing the \textit{global} context over the whole data space. Therefore, AAF can be viewed as a specific case of our contrastive loss, and additionally considering AAF  does not bring any performance improvement. For other losses,  our contrastive loss are complementary to them (global \textit{vs.}~local, pairwise \textit{vs.}~higher-order) and thus enables further performance uplifting. This suggests that designing a higher-order, global loss for semantic segmentation is a promising direction.

\section{Additional Quantitative Result}\label{sec:city-val}

Table~\ref{table:city-val} provides comparison results with representative approaches on Cityscapes \texttt{val}~\cite{cordts2016cityscapes} in terms of mIoU and training speed. We train our models  on Cityscapes \texttt{train} for $80,\!000$ iterations with a mini-batch size of $8$. We find that, by equipping with cross-image pixel contrast, the performance of baseline models enjoy consistently improvements (\textbf{1.2}/\textbf{1.1}/\textbf{0.8} points gain over DeepLabV3, HRNetV2 and OCR, respectively). In addition, the contrastive loss computation  brings negligible training speed decrease, and does not incur any extra overhead during inference.

\begin{table}[t]
	\centering
	\small
	\resizebox{0.49\textwidth}{!}{
		\setlength\tabcolsep{4pt}
		\renewcommand\arraystretch{1.0}
		\begin{tabular}{r||c|c|c}
			\hline\thickhline
			\rowcolor{mygray}
			Model & Backbone & sec./iter. & mIoU (\%) \\ \hline\hline
			SegSort$_{19}$~\cite{hwang2019segsort} & D-ResNet-101  &-& 78.2 \\
			AAF$_{18}$~\cite{ke2018adaptive} & D-ResNet-101 &-& 79.2 \\
			DeepLabV3+$_{18}$~\cite{chen2018encoder} & D-Xception-71 &-& 79.6 \\
			PSPNet$_{17}$~\cite{zhao2017pyramid} & D-ResNet-101 &-& 79.7 \\
			Auto-DeepLab-L$_{19}$~\cite{liu2019auto} & - & - & 80.3 \\
			HANet$_{20}$~\cite{choi2020cars} & D-ResNet-101 &-& 80.3 \\
			SpyGR$_{20}$~\cite{li2020spatial} & D-ResNet-101 & -&80.5 \\
			ACF$_{19}$~\cite{zhang2019acfnet} & D-ResNet-101 & - & 81.5 \\
			\hline\hline
			
			DeepLabV3$_{17}$~\cite{chen2017rethinking} & & 1.18 & 78.5 \\
			DeepLabV3$+~\textbf{\texttt{Ours}}$& \multirow{-2}{*}{D-ResNet-101}& 1.37 & \textbf{79.7} \color{mygreen}{(\textbf{+1.2})} \\ \hline

			HRNetV2$_{20}$~\cite{wang2020deep} &  & 1.67  & 81.1 \\
			HRNetV2$+~\textbf{\texttt{Ours}}$& \multirow{-2}{*}{HRNetV2-W48} & 1.87 & \textbf{82.2} \color{mygreen}{(\textbf{+1.1})} \\ \hline
			
			OCR$_{20}$~\cite{yuan2020object} &  & 1.29 & 80.6 \\
			OCR$+~\textbf{\texttt{Ours}}$& \multirow{-2}{*}{D-ResNet-101}& 1.41 & \textbf{81.2} \color{mygreen}{(\textbf{+0.6})}  \\
			
			OCR$_{20}$~\cite{yuan2020object} &  & 1.75 & 81.6 \\
			OCR$+~\textbf{\texttt{Ours}}$& \multirow{-2}{*}{HRNetV2-W48} & 1.90 & \textbf{82.4} \color{mygreen}{(\textbf{+0.8})} \\
			
			\hline
		\end{tabular}
	}
	\captionsetup{font=small}
	\caption{\small\textbf{Quantitative semantic segmentation results} on Cityscapes \texttt{val}\!~\cite{cordts2016cityscapes}. D-ResNet-101 = Dilated-ResNet-101. D-Xception-71 = Dilated-Xception-71.  See \S\ref{sec:city-val} for more details.}
	\label{table:city-val}
	\vspace{-12pt}
\end{table}

\section{Additional Qualitative Result}\label{sec:qualitative}
\begin{figure*}[t]
	\begin{center}
		\includegraphics[width=\linewidth]{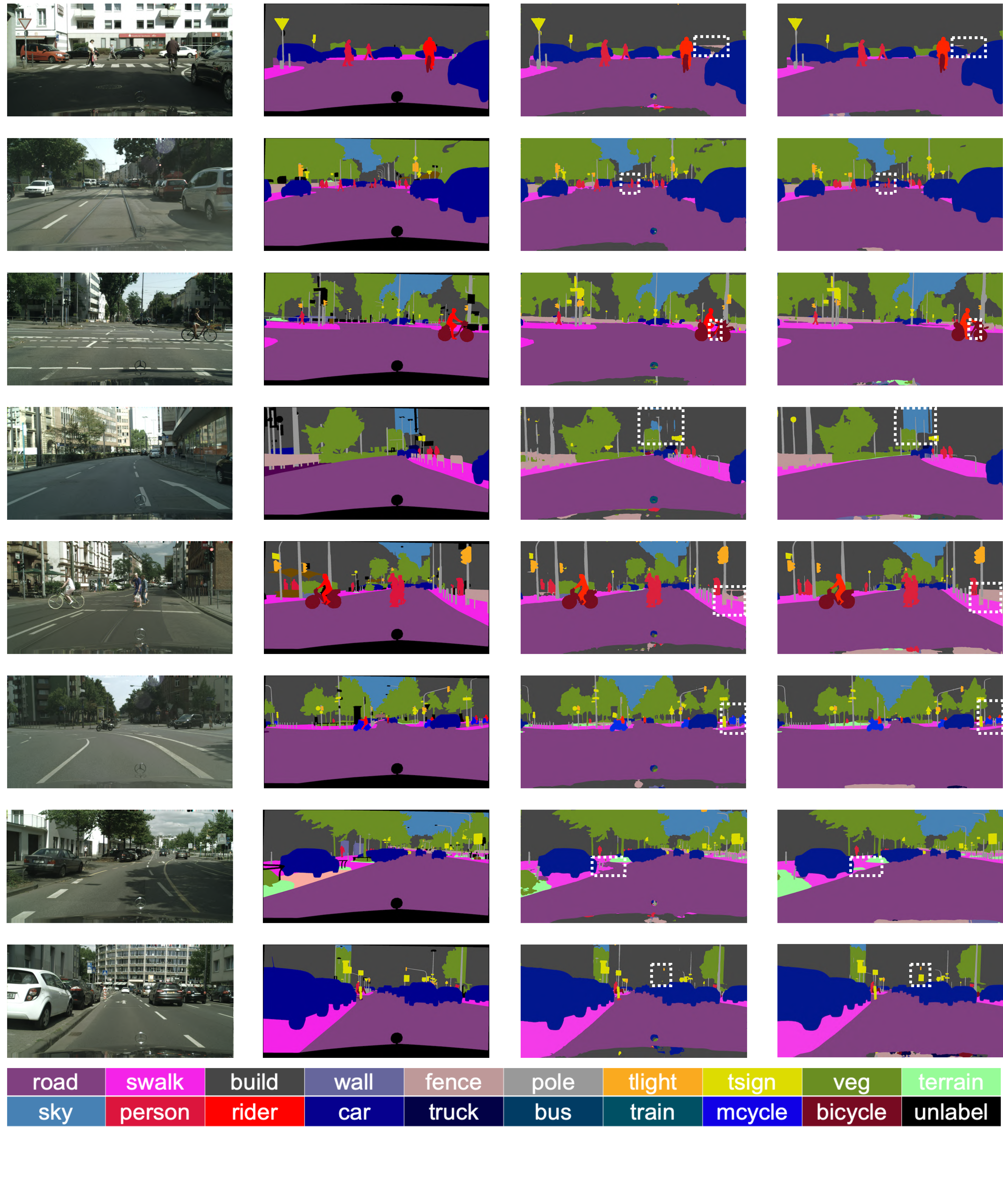}
	\end{center}
	\vspace{-10pt}
	\captionsetup{font=small}
	\caption{\small\textbf{Qualitative semantic segmentation results} on Cityscapes \texttt{val}~\cite{cordts2016cityscapes}. From left to right: input images, ground-truths, results of HRNetV2~\cite{wang2020deep}, results of HRNetV2$+$Ours. The improved regions are marked by white dashed boxes.}
	\label{fig:1}
\end{figure*}

\begin{figure*}[t]
	\begin{center}
		\includegraphics[width=\linewidth]{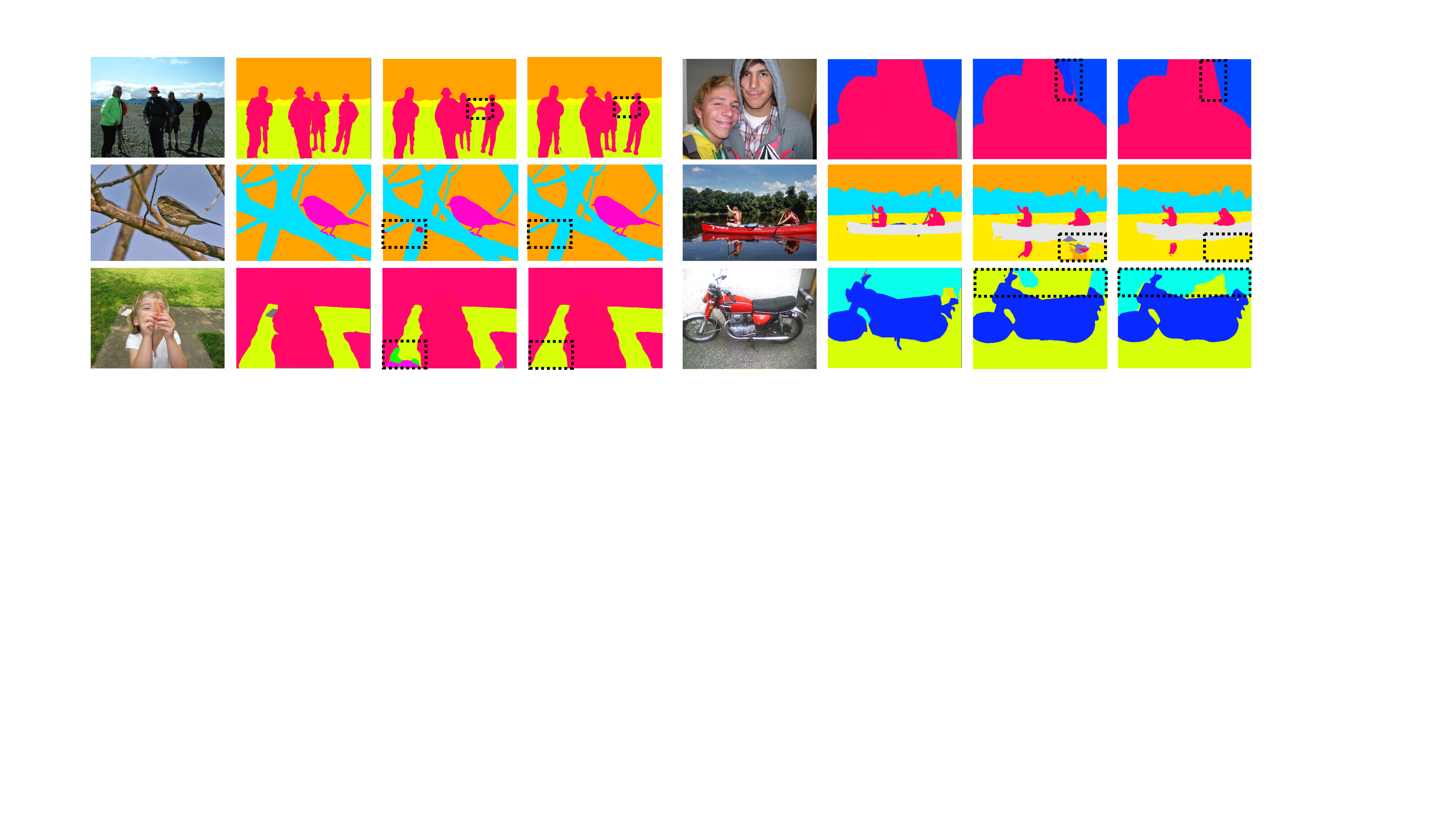}
	\end{center}
	\captionsetup{font=small}
	\caption{\small\textbf{Qualitative semantic segmentation results} on PASCAL-Context \texttt{test}~\cite{mottaghi2014role}. From left to right: input images, ground-truths, results of HRNetV2~\cite{wang2020deep}, results of HRNetV2$+$Ours. The improved regions are marked by black dashed boxes.}
	\label{fig:2}
\end{figure*}

\begin{figure*}[t]
	\begin{center}
		\includegraphics[width=\linewidth]{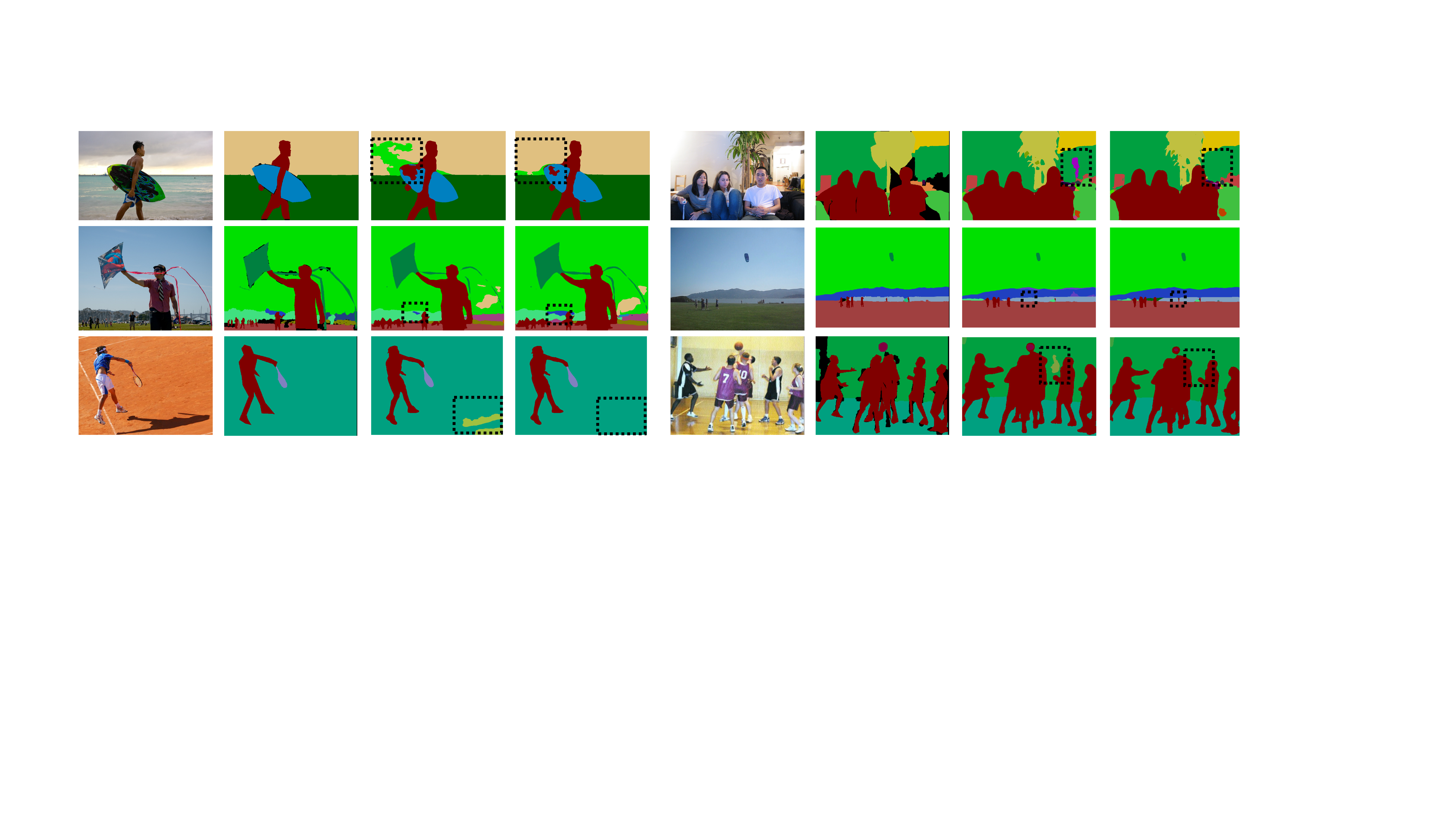}
	\end{center}
	\captionsetup{font=small}
	\caption{\small\textbf{Qualitative semantic segmentation results} on COCO-Stuff \texttt{test}~\cite{caesar2018coco}. From left to right: input images, ground-truths, results of HRNetV2~\cite{wang2020deep}, results of HRNetV2$+$Ours. The improved regions are marked by black dashed boxes.}
	\label{fig:3}
\end{figure*}

\begin{figure*}[t]
	\begin{center}
		\includegraphics[width=\linewidth]{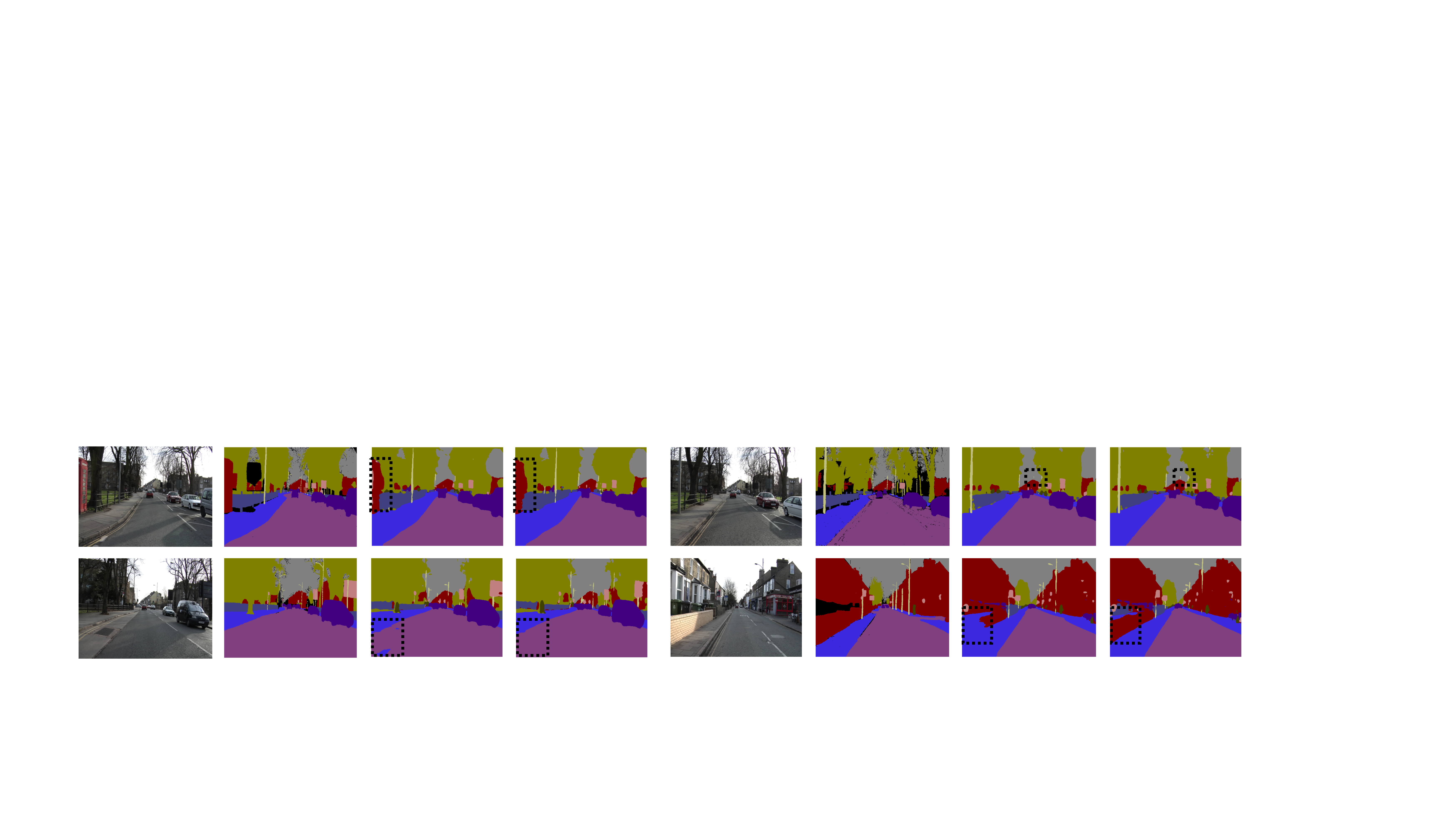}
	\end{center}
	\captionsetup{font=small}
	\caption{\small\textbf{Qualitative semantic segmentation results} on CamVid \texttt{test}~\cite{brostow2009semantic}. From left to right: input images, ground-truths, results of HRNetV2~\cite{wang2020deep}, results of HRNetV2$+$Ours. The improved regions are marked by black dashed boxes.}
	\label{fig:4}
\end{figure*}

We provide additional qualitative improvements of HRNetV2$+$Ours over HRNetV2~\cite{wang2020deep} on four benchmarks, including Cityscapes \texttt{val}~\cite{cordts2016cityscapes} in~\figref{fig:1}, PASCAL-Context \texttt{test}~\cite{mottaghi2014role} in~\figref{fig:2}, COCO-Stuff \texttt{test}~\cite{caesar2018coco} in~\figref{fig:3}, and CamVid \texttt{test}~\cite{brostow2009semantic} in~\figref{fig:4}. The improved regions are marked by dashed boxes. As can be seen, our approach is able to produce great improvements on those hard regions, \eg, small objects, cluttered background.

\mbox{}
\clearpage

{\small
\bibliographystyle{ieee_fullname}
\bibliography{egbib}
}

\end{document}